\documentclass[11pt]{article}
\usepackage{amssymb}
\usepackage{multirow}
\usepackage{amsmath}
\usepackage{amsfonts}
 \usepackage{pifont}
\usepackage{amsmath,amsfonts,amssymb,theorem,graphicx,listings,txfonts}
\usepackage{graphics}
 \usepackage{caption2}
\usepackage{flafter}
 \usepackage{indentfirst}
\usepackage{epsfig}
  \usepackage{mathrsfs}
  \usepackage{float}
\usepackage{subfigure}
 \usepackage{cite}
\newtheorem{theorem}{Theorem}[section]

\newtheorem{definition}[theorem]{Definition}

\theoremstyle{example}
\newtheorem{example}[theorem]{Example}
\theoremstyle{programme}

\theoremstyle{property}

\theoremstyle{problem}

\renewcommand{\arraystretch}{1}

\title{Conflict Analysis for Pythagorean Fuzzy Information Systems with Group Decision Making}

\author
{Guangming Lang$^{1,2,3}$  
\thanks{Corresponding author.\quad
langguangming1984@tongji.edu.cn(G.M.Lang)
\newline\mbox{}\hspace{0.55cm}}\\
\small {$^{1}$ Department of Computer Science and Technology, Tongji University}\\
\small {Shanghai 201804, P.R. China}\\
\small {$^{2}$ School of Mathematics and Statistics, Changsha University of Science and Technology}\\
\small {Changsha, Hunan 410114, P.R. China}\\
\small {$^{3}$ The Key Laboratory of Embedded System and Service Computing, Ministry of Education, Tongji University}\\
\small {Shanghai 201804, P.R. China}\\
}
\date{}

\begin{document}
\maketitle \baselineskip=17pt
\begin{center}
\begin{quote}
{{\bf Abstract.}
Pythagorean fuzzy sets provide stronger ability than intuitionistic fuzzy sets to model uncertainty information and knowledge, but little effort has been paid to conflict analysis of Pythagorean fuzzy information systems. In this paper, we present three types of positive, central, and negative alliances with different thresholds, and employ examples to illustrate how to construct the positive, central, and negative alliances. Then we study conflict analysis of Pythagorean fuzzy information systems based on Bayesian minimum risk theory. Finally, we investigate group conflict analysis of Pythagorean fuzzy information systems based on Bayesian minimum risk theory.

\\}
\end{quote}
\end{center}
\renewcommand{\thesection}{\arabic{section}}

\section{Introduction}

Pythagorean fuzzy sets(PFSs), as a generalization of intuitionistic fuzzy sets(IFSs), are characterized by a membership degree and a non-membership degree satisfying the condition that the square sum of its membership degree and non-membership degree is equal to or less than $1$, and they have more powerful ability than IFSs to model the uncertain information in decision making problems. So far, much effort\cite{Atanassov,Bustince,Peng,Reformat,Ren,Yager,Yager1,Yang1,Zhang,Zhang1,Zhang2} have paid to investigations of Pythagorean fuzzy sets. For example, Beliakov et al.\cite{Beliakov} provided the averaging aggregation functions for preferences expressed as Pythagorean membership grades and fuzzy orthopairs. Bustince et al.\cite{Bustince} investigated a historical account of types of fuzzy sets and discussed their relationships. Peng\cite{Peng} developed a Pythagorean fuzzy superiority and inferiority ranking method to solve uncertainty multiple attribute group decision making problem. Based on Pythagorean fuzzy sets, Reformat et al.\cite{Reformat} proposed a novel collaborative-based recommender system that provides a user with the ability to control a process of constructing a list of suggested items. Ren et al.\cite{Ren} extended the TODIM approach to solve the MCDM problems with Pythagorean fuzzy information and analyzed how the risk attitudes of the decision makers exert the influence on the results of MCDM under uncertainty. Yager\cite{Yager} introduced a variety of aggregation operations for Pythagorean fuzzy subsets and studied multi-criteria decision making where the criteria satisfaction are expressed using Pythagorean membership grades. Zhang\cite{Zhang} presented a hierarchical QUALIFLEX approach with the closeness index-based ranking methods for multi-criteria Pythagorean fuzzy decision analysis. Zhang et al.\cite{Zhang1} defined some novel operational laws of PFSs and proposed an extended technique for order preference by similarity to ideal solution method to deal effectively with the multi-criteria decision-making problems with PFSs. Zhang et al.\cite{Zhang2} introduced the models of Pythagorean fuzzy rough set over two universes and Pythagorean fuzzy multi-granulation rough set over two universes.

Many scholars\cite{Jiang,Liu,Pawlak,Pawlak1,Pawlak2,Silva,Sun,Sun1,Yang,Yu,
Deja1,Deja2,Deja3,Deja4,Fraser,Liu1,Lin,Maeda,Nguyen,Ramanna,Ramanna1,Ramanna3,
Ramanna4,Saaty,Skowron,Skowron1,Zhu1,Zhu2} focused on conflict analysis of information systems, and improved the relationship between the two sides of a conflict by finding the essence of the conflict issue. For example, Deja\cite{Deja2} examined nature of conflicts as we are formally defining the conflict situation model. Pawlak\cite{Pawlak1} initially considered the auxiliary functions and distance functions and offered deeper insight into the structure of conflicts and enables the analysis of relationships between parties and the issues being debated. Silva et al.\cite{Silva} presented a multicriteria approach for analysis of conflicts in evidence theory. Ramanna et al.\cite{Ramanna1} studied how to model a combination of complex situations among agents where there are disagreements leading to a conflict situation. Sun et al.\cite{Sun1} subsequently proposed a conflict analysis decision model and developed a matrix approach for conflict analysis based on rough set theory over two universes.
Skowron et al.\cite{Skowron} explained the nature of conflict and defined the conflict situation
model in a way to encapsulate the conflict components in a clear manner.
Yang et al.\cite{Yang} investigated evidence conflict and belief convergence based on the analysis of the degree of coherence between two sources of evidence and illustrated the stochastic interpretation for basic probability assignment. Yu et al.\cite{Yu} provided the supporting probability distance to characterize the differences among bodies of evidence and gave a new combination rule for the combination of the conflicting evidence. Zhu et al.\cite{Zhu2} studied the problems of conflicts of interest in database access security using granular computing based on covering rough set theory.

In practical situations, many agents express their opinions on some issues by Pythagorean fuzzy sets, which are more precise than intuitionistic fuzzy sets and other types of fuzzy sets. For example, an agent expresses his preference about the degree of an issue, and he may give the degree to support this issue as $\frac{\sqrt{3}}{2}$, and the degree to  nonsupport this issue as $\frac{1}{2}$. Thus we have $(\frac{\sqrt{3}}{2})^{2}+(\frac{1}{2})^{2}=1$ and $\frac{\sqrt{3}}{2}+\frac{1}{2}>1$. We see Pythagorean fuzzy sets are more effective than intuitionistic fuzzy sets for expressing uncertain information in practical situations.
Although there are many Pythagorean fuzzy information systems for conflicts, few researches focus on conflict analysis of Pythagorean fuzzy information systems. Furthermore, due to the characteristic of data collection actually, there are many dynamic Pythagorean fuzzy information systems with variations of object sets, attribute sets, and attribute value sets, and non-incremental approaches are time-consuming for conflict analysis of Pythagorean fuzzy information systems. Therefore, we should provide effective approaches for conflict analysis of dynamic Pythagorean fuzzy information systems.

The contributions of this paper are as follows. Firstly, we provide the concept of
Pythagorean fuzzy information system and employ an example to illustrate the difference between information systems and Pythagorean fuzzy information systems. We present the concepts of Pythagorean matrix, Pythagorean closeness index matrix, closeness index, comprehensive Pythagorean closeness index. We also provide the concepts of positive alliance, central alliance, and negative alliance. Especially, We present the concepts of maximum positive alliance, central alliance, and negative alliance. We propose an algorithm for conflict analysis of Pythagorean fuzzy information systems and employ an example to illustrate how to conduct conflict analysis of Pythagorean fuzzy information systems. Secondly, we provide the concept of dynamic Pythagorean fuzzy information systems when adding an object, and discuss the relationship between the original Pythagorean fuzzy information systems and dynamic Pythagorean fuzzy information systems. We propose an incremental algorithm for conflict analysis of Pythagorean fuzzy information system when adding an object. Thirdly, we provide the concept of dynamic Pythagorean fuzzy information systems when deleting an object, and discuss the relationship between the original Pythagorean fuzzy information systems and dynamic Pythagorean fuzzy information systems. We propose an incremental algorithm for conflict analysis of Pythagorean fuzzy information system when deleting an object.

The rest of this paper is shown as follows. Section 2 reviews the basic concepts of Pythagorean fuzzy sets. Section 3 provides three types of positive, central, and negative alliances with different thresholds. Section 4 investigates conflict analysis of Pythagorean fuzzy information systems based on Bayesian minimum risk theory. Section 5 focuses group conflict analysis of Pythagorean fuzzy information systems based on Bayesian minimum risk theory. The conclusion is given in Section 6.

\section{Preliminaries}

In this section, we review the related concepts of Pythagorean fuzzy sets.

\begin{definition}\cite{Yager}
Let $U$ be an arbitrary non-empty set, and a Pythagorean fuzzy set(PFS) $P$ is a mathematical object of the form as follows:
\begin{eqnarray*}
P=\{<x,P(\mu_{P}(x),\nu_{P}(x))>|x\in U\},
\end{eqnarray*}
where $\mu_{P}(x),\nu_{P}(x):U\rightarrow [0,1]$ such as $\mu^{2}_{P}(x)+\nu^{2}_{P}(x)\leq 1$, for every $x\in U$, $\mu_{P}(x)$ and $\nu_{P}(x)$ denote the membership degree and the non-membership degree of the element $x$ to $U$ in $P$, respectively.
\end{definition}

For convenience, we denote the Pythagorean fuzzy number (PFN) and the hesitant degree as $\gamma=P(\mu_{\gamma},\nu_{\gamma})$, and $\pi_{\gamma}=\sqrt{1-\mu^{2}_{\gamma}-\nu^{2}_{\gamma}}$, respectively. Moreover, as a generalization of Intuitionistic fuzzy sets, Pythagorean fuzzy sets are powerful for describing imprecise information in practice.

\begin{definition}\cite{Yager}
Let $\gamma_{1}=P(\mu_{\gamma_{1}},\nu_{\gamma_{1}})$ and $\gamma_{2}=P(\mu_{\gamma_{2}},\nu_{\gamma_{2}})$ be Pythagorean fuzzy numbers, a nature quasi-ordering on the PFNs is defined as follows:
$$\gamma_{1}\geq\gamma_{2} \text{ if and only if } \mu_{\gamma_{1}}\geq \mu_{\gamma_{2}} \text{ and } \nu_{\gamma_{1}}\leq \nu_{\gamma_{2}}.$$
\end{definition}

Moreover, Yager also provided summation and multiplication operations for Pythagorean fuzzy numbers as follows.

$(1)$ $\gamma_{1}\oplus\gamma_{2}=P(\sqrt{\mu^{2}_{\gamma_{1}}+\mu^{2}_{\gamma_{2}}-\mu^{2}_{\gamma_{1}}\ast\mu^{2}_{\gamma_{2}}},
\nu_{\gamma_{1}}\ast\nu_{\gamma_{2}})$;

$(2)$ $k\gamma=P(\sqrt{1-(1-\mu^{2}_{\gamma})^{k}},\nu^{k}_{\gamma}).$

Generally speaking, it is natural to discern Pythagorean fuzzy numbers by Definition 2.2. Actually, it does not work in some situations. For example, if $\gamma_{1}=P(\frac{\sqrt{5}}{3},\frac{\sqrt{2}}{3})$ and $\gamma_{2}=P(\frac{\sqrt{2}}{3},\frac{1}{3})$, then $\frac{\sqrt{5}}{3}>\frac{\sqrt{2}}{3}$ and $\frac{\sqrt{2}}{3}>\frac{1}{3}$. Thus we can not compare $\gamma_{1}$ and $\gamma_{2}$ by Definition 2.2. Subsequently, Yager proposed the score function for Pythagorean fuzzy numbers as follows.

\begin{definition}
Let $\gamma=P(\mu_{\gamma},\nu_{\gamma})$ be a Pythagorean fuzzy number. Then the score function of $\gamma$ is defined as follows:
$$S(\gamma)=\mu^{2}_{\gamma}-\nu^{2}_{\gamma}.$$
\end{definition}

By Definition 2.3, we have $-1\leq S(\gamma)\leq 1$ for the Pythagorean fuzzy number $\gamma$. Especially, the score function is more effective for comparing
some Pythagorean fuzzy numbers. Moreover, Yager introduced the comparison law for Pythagorean fuzzy numbers as follows.

\begin{definition}
Let $\gamma_{1}=P(\mu_{\gamma_{1}},\nu_{\gamma_{1}})$ and $\gamma_{2}=P(\mu_{\gamma_{2}},\nu_{\gamma_{2}})$ be Pythagorean fuzzy numbers. Then we have

$(1)$ If $S(\gamma_{1})>S(\gamma_{2}),$ then $\gamma_{1}$ is bigger than $\gamma_{2}$, denoted by $\gamma_{1}\succ_{S}\gamma_{2}$;

$(2)$ If $S(\gamma_{1})<S(\gamma_{2}),$ then $\gamma_{1}$ is smaller than $\gamma_{2}$, denoted by $\gamma_{1}\prec_{S}\gamma_{2}$;

$(3)$ If $S(\gamma_{1})=S(\gamma_{2}),$ then $\gamma_{1}$ is equal to $\gamma_{2}$, denoted by $\gamma_{1}\sim_{S}\gamma_{2}$.
\end{definition}

By Definition 2.3, we have $S(\gamma_{1})=S(P(\frac{\sqrt{5}}{3},\frac{\sqrt{2}}{3}))=\frac{3}{9}$ and $S(\gamma_{2})=S(P(\frac{\sqrt{2}}{3},\frac{1}{3}))=\frac{1}{9}$. Therefore, we have $S(\gamma_{1})<S(\gamma_{2})$. Moreover, the score function may fail to compare Pythagorean fuzzy numbers in practical situations as follows.

\begin{example}
Let $\gamma_{1}=P(\frac{\sqrt{5}}{3},\frac{2}{3})$ and $\gamma_{2}=P(\frac{2}{3},\frac{\sqrt{3}}{3})$. Then we have
\begin{eqnarray*}
S(\gamma_{1})=(\frac{\sqrt{5}}{3})^{2}-(\frac{2}{3})^{2}=\frac{1}{9} \text{ and }
S(\gamma_{2})=(\frac{2}{3})^{2}-(\frac{\sqrt{3}}{3})^{2}=\frac{1}{9}.
\end{eqnarray*}
\end{example}

Apparently, we get $S(\gamma_{1})=S(\gamma_{2})$. But there are some differences between $\gamma_{1}$ and $\gamma_{2}$. Thus the score function fails to compare Pythagorean fuzzy numbers in this situation. Therefore, we should provide effective measures for Pythagorean fuzzy numbers.

\begin{definition}\cite{Zhang1}
Let $\gamma_{1}=P(\mu_{\gamma_{1}},\nu_{\gamma_{1}})$ and $\gamma_{2}=P(\mu_{\gamma_{2}},\nu_{\gamma_{2}})$ be Pythagorean fuzzy numbers. Then the Euclidean distance between $\gamma_{1}$ and $\gamma_{2}$ is defined as follows:
\begin{eqnarray*}
d(\gamma_{1},\gamma_{2})=\frac{1}{2}(|\mu^{2}_{\gamma_{1}}-\mu^{2}_{\gamma_{2}}|+|\nu^{2}_{\gamma_{1}}
-\nu^{2}_{\gamma_{2}}|+|\pi^{2}_{\gamma_{1}}
-\pi^{2}_{\gamma_{2}}|).
\end{eqnarray*}
\end{definition}

Especially, we obtain the Euclidean distance between the Pythagorean fuzzy number $P(\mu_{\gamma},\nu_{\gamma})$ and the positive ideal PFN $\gamma^{+}=P(1,0)$ as follows:
\begin{eqnarray*}
d(\gamma,\gamma^{+})=\frac{1}{2}(1-\mu^{2}_{\gamma}+\nu^{2}_{\gamma}
+\pi^{2}_{\gamma})=1-\mu^{2}_{\gamma},
\end{eqnarray*}
and the Euclidean distance between the Pythagorean fuzzy number $P(\mu_{\gamma},\nu_{\gamma})$ and the negative ideal PFN $\gamma^{-}=P(0,1)$ as follows:
\begin{eqnarray*}
d(\gamma,\gamma^{-})=\frac{1}{2}(1-\nu^{2}_{\gamma}+\mu^{2}_{\gamma}
+\pi^{2}_{\gamma})=1-\nu^{2}_{\gamma}.
\end{eqnarray*}

By Definition 2.6, we provide the closeness index for Pythagorean fuzzy numbers as follows.

\begin{definition}\cite{Zhang1}
Let $\gamma=P(\mu_{\gamma},\nu_{\gamma})$ be a Pythagorean fuzzy number, $\gamma^{+}=P(1,0)$,  and $\gamma^{-}=P(0,1)$. Then the closeness index of $\gamma$ is defined as follows:
\begin{eqnarray*}
\mathscr{P}(\gamma)=\frac{d(\gamma,\gamma^{-})}{d(\gamma,\gamma^{+})+d(\gamma,\gamma^{-})}
=\frac{1-\nu^{2}_{\gamma}}{2-\mu^{2}_{\gamma}-\nu^{2}_{\gamma}}.
\end{eqnarray*}
\end{definition}

By Definition 2.7, we see that the closeness index $\mathscr{P}(\gamma)$ of $\gamma$ is constructed based on the Euclidean distance between the Pythagorean fuzzy number $P(\mu_{\gamma},\nu_{\gamma})$ and the positive ideal PFN $\gamma^{+}$ and the Euclidean distance between the Pythagorean fuzzy number $P(\mu_{\gamma},\nu_{\gamma})$ and the negative ideal PFN $\gamma^{-}$. Especially, we have $0\leq \mathscr{P}(\gamma)\leq 1$ for the Pythagorean fuzzy number $\gamma$.

\begin{definition}
Let $\gamma_{1}=P(\mu_{\gamma_{1}},\nu_{\gamma_{1}})$ and $\gamma_{2}=P(\mu_{\gamma_{2}},\nu_{\gamma_{2}})$ be Pythagorean fuzzy numbers. Then we have

$(1)$ If $\mathscr{P}(\gamma_{1})>\mathscr{P}(\gamma_{2}),$ then $\gamma_{1}$ is bigger than $\gamma_{2}$, denoted by $\gamma_{1}\succ_{\mathscr{P}}\gamma_{2}$;

$(2)$ If $\mathscr{P}(\gamma_{1})<\mathscr{P}(\gamma_{2}),$ then $\gamma_{1}$ is smaller than $\gamma_{2}$, denoted by $\gamma_{1}\prec_{\mathscr{P}}\gamma_{2}$;

$(3)$ If $\mathscr{P}(\gamma_{1})=\mathscr{P}(\gamma_{2}),$ then $\gamma_{1}$ is equal to $\gamma_{2}$, denoted by $\gamma_{1}\sim_{\mathscr{P}}\gamma_{2}$.
\end{definition}

\begin{example}\cite{Yager}(Continuation from Example 2.5)
By Definition 2.7, we have
$$\mathscr{P}(\gamma_{1})=\frac{1-\frac{4}{9}}{2-\frac{5}{9}-\frac{4}{9}}=\frac{5}{9} \text{ and }
\mathscr{P}(\gamma_{2})=\frac{1-\frac{1}{3}}{2-\frac{4}{9}-\frac{3}{9}}=\frac{6}{11}.$$
Therefore, by Definition 2.8, we get $\mathscr{P}(\gamma_{1})>\mathscr{P}(\gamma_{2})$.
\end{example}

\section{Conflict analysis for Pythagorean fuzzy information systems}

In this section, we investigate conflict analysis for Pythagorean fuzzy information systems.

\begin{definition}
A Pythagorean fuzzy information system is a 4-tuple $S=(U, A, V, f),$ where
$U=\{x_{1},x_{2},...,\\x_{n}\}$ is a finite set of objects, $A=\{c_{1},c_{2},...,c_{m}\}$ is a
finite set of attributes, $V=\{V_{a}\mid a\in A\}$, where $V_{a}$ is the
set of attribute values on $a$, all attribute values are Pythagorean fuzzy numbers, and $f$ is a
function from $U\times A$ into $V$.
\end{definition}

Pythagorean fuzzy information systems, as a generalization of Pawlak's information systems, represent all available information and knowledge, where objects are measured by using a finite number of attributes and attribute values are PFNs. Furthermore, Pythagorean fuzzy information systems can provide more information than Intuitionistic fuzzy information systems, which are powerful for dealing with uncertain information and knowledge.

\begin{example} Table 1 depicts a Pythagorean fuzzy information system for
conflicts, where $x_{1}, x_{2}, x_{3}, x_{4}, x_{5}$, and $x_{6}$ denote six agents, $c_{1}, c_{2},
c_{3}, c_{4}, c_{5}$, and $c_{6}$ denote six issues. For example, we have $c_{1}(x_{1})=P(\mu_{P}(x_{1}),\nu_{P}(x_{1}))=P(1.0,0.0)$, where $\mu_{P}(x_{1})=1.0$ denotes the support degree of the agent $x_{1}$ to the issue $c_{1}$, and $\nu_{P}(x_{1})=0.0$ denotes the opposing degree of the agent $x_{1}$ to the issue $c_{1}$; we have $c_{5}(x_{6})=P(\mu_{P}(x_{1}),\nu_{P}(x_{1}))=P(0.8,0.4)$, where $\mu_{P}(x_{1})=0.8$ denotes the support degree of the agent $x_{6}$ to the issue $c_{5}$, and $\nu_{P}(x_{6})=0.4$ denotes the opposing degree of the agent $x_{6}$ to the issue $c_{5}$.
\begin{table}[H]\renewcommand{\arraystretch}{1.1}
\caption{The Pythagorean fuzzy information system for conflicts.}
\tabcolsep0.25in
\begin{tabular}{ c c c c c c}
\hline $U$ &$c_{1}$ & $c_{2}$& $c_{3}$& $c_{4}$& $c_{5}$\\\hline
$x_{1}$& $P(1.0,0.0)$ & $P(0.9,0.3)$ & $P(0.8,0.2)$ & $P(0.9,0.1)$ & $P(0.9,0.2)$ \\
$x_{2}$& $P(0.9,0.1)$ & $P(0.5,0.5)$ & $P(0.1,0.9)$ & $P(0.3,0.8)$ & $P(0.1,0.9)$ \\
$x_{3}$& $P(0.1,0.9)$ & $P(0.1,0.9)$ & $P(0.2,0.8)$ & $P(0.1,0.9)$ & $P(0.5,0.5)$ \\
$x_{4}$& $P(0.5,0.5)$ & $P(0.1,0.9)$ & $P(0.3,0.7)$ & $P(0.5,0.5)$ & $P(0.1,0.9)$ \\
$x_{5}$& $P(0.9,0.2)$ & $P(0.4,0.6)$ & $P(0.1,0.9)$ & $P(0.1,0.9)$ & $P(0.3,0.9)$ \\
$x_{6}$& $P(0.0,1.0)$ & $P(0.9,0.1)$ & $P(0.2,0.9)$ & $P(0.5,0.5)$ & $P(0.8,0.4)$ \\
\hline
\end{tabular}
\end{table}
\end{example}

\begin{definition}
Let $S=(U, A, V, f)$ be a Pythagorean fuzzy information system. Then the Pythagorean matrix $M(S)$ is defined as follows:
\begin{eqnarray*}
M(S)=\left[
  \begin{array}{cccccc}
    P(\mu_{11},\nu_{11}) & P(\mu_{12},\nu_{12}) & . & . & . &P(\mu_{1m},\nu_{1m}) \\
    P(\mu_{21},\nu_{21}) & P(\mu_{22},\nu_{22}) & . & . & . &P(\mu_{2m},\nu_{2m}) \\
    . & . & . & . & . & . \\
    . & . & . & . & . & . \\
    . & . & . & . & . & . \\
    P(\mu_{n1},\nu_{n1}) & P(\mu_{n2},\nu_{n2}) & . & . & . &P(\mu_{nm},\nu_{nm}) \\
  \end{array}
\right].
\end{eqnarray*}
\end{definition}

By Definition 3.3, we transform a Pythagorean fuzzy information system into a matrix for conflict analysis. Furthermore, we employ an example to illustrate how to transform Pythagorean fuzzy information systems into matrices as follows.

\begin{example}(Continuation from Example 3.2)
By Definition 3.3, we have the Pythagorean matrix $M(S)$ as follows:
\begin{eqnarray*}
M(S)=\left[
\begin{array}{ccccc}
P(1.0,0.0) & P(0.9,0.3) & P(0.8,0.2) & P(0.9,0.1) & P(0.9,0.2) \\
P(0.9,0.1) & P(0.5,0.5) & P(0.1,0.9) & P(0.3,0.8) & P(0.1,0.9) \\
P(0.1,0.9) & P(0.1,0.9) & P(0.2,0.8) & P(0.1,0.9) & P(0.5,0.5) \\
P(0.5,0.5) & P(0.1,0.9) & P(0.3,0.7) & P(0.5,0.5) & P(0.1,0.9) \\
P(0.9,0.2) & P(0.4,0.6) & P(0.1,0.9) & P(0.1,0.9) & P(0.3,0.9) \\
P(0.0,1.0) & P(0.9,0.1) & P(0.2,0.9) & P(0.5,0.5) & P(0.8,0.4) \\
\end{array}
\right].
\end{eqnarray*}
\end{example}

\begin{definition}\cite{Yager}
Let $\mathscr{P}=\{\gamma_{i}|\gamma_{i}=P(\mu_{\gamma_{i}},\nu_{\gamma_{i}}),
i=1,2,...,m\}$ be a collection of PFNs, and $\mathscr{K}=\{k_{1},k_{2},...,k_{m}\}$ be the weight vector of $\gamma_{i} $ $(i=1,2,...,m)$, where $k_{i}$ indicates the importance degree of $\gamma_{i}$, satisfying $k_{i}\geq 0$ $(i=1,2,...,m)$ and $\Sigma_{i=1}^{m}k_{i}=1$. Then the Pythagorean fuzzy weighted averaging operator $\mathscr{R}$: $\Theta^{m}\rightarrow \Theta$ is defined as follows:
\begin{eqnarray*}
\mathscr{R}(\gamma_{1},\gamma_{2},...,\gamma_{m})=P(\Sigma_{i=1}^{m}k_{i}\mu_{\gamma_{i}},\Sigma_{i=1}^{m}k_{i}\nu_{\gamma_{i}}).
\end{eqnarray*}
\end{definition}

By Definition 3.5, we aggregate a collection of Pythagorean fuzzy numbers $\{\gamma_{i}|\gamma_{i}=P(\mu_{\gamma_{i}},\nu_{\gamma_{i}}),
i=1,2,...,m\}$ into a Pythagorean fuzzy number $\mathscr{R}(\gamma_{1},\gamma_{2},...,\gamma_{m})$ with the weight vector, and $\mathscr{R}(\gamma_{1},\gamma_{2},...,\gamma_{m})$ is a Pythagorean fuzzy number. For simplicity,  we denote $\mathscr{R}(c_{1}(x),c_{2}(x),,...,c_{m}(x))$ as $\mathscr{R}(x)$ in the following discussion. Moreover, we provide three concepts of the positive, central, and negative alliances using the Pythagorean fuzzy weighted averaging operator as follows.

\begin{definition}
Let $S=(U, A, V, f)$ be a Pythagorean fuzzy information system, $\gamma^{\circ}$ and $\gamma_{\circ}$ are PFSs. Then the positive, central, and
negative alliances are defined as follows:
\begin{eqnarray*}
POA_{(\gamma^{\circ},\gamma_{\circ})}(U)&=&\{x\in U\mid \mathscr{R}(x)\geq \gamma^{\circ}\};\\
CTA_{(\gamma^{\circ},\gamma_{\circ})}(U)&=&\{x\in U\mid \gamma_{\circ}< \mathscr{R}(x)<\gamma^{\circ}\};\\
NEA_{(\gamma^{\circ},\gamma_{\circ})}(U)&=&\{x\in U\mid \mathscr{R}(x)\leq \gamma_{\circ}\}.
\end{eqnarray*}
\end{definition}

By Definition 3.6, we get the positive, central, and negative alliances using the Pythagorean fuzzy numbers $\gamma^{\circ}$ and $\gamma_{\circ}$. Especially, we have $POA_{(\gamma^{\circ},\gamma_{\circ})}(U)\cup
CTA_{(\gamma^{\circ},\gamma_{\circ})}(U)\cup
NEA_{(\gamma^{\circ},\gamma_{\circ})}(U)\subseteq U$. Furthermore, although we can compare Pythagorean fuzzy numbers directly by Definition 2.2, it doesn't work in some situations shown as examples after Definition 2.2.

\begin{example}(Continuation from Example 3.4) Taking $w_{1}=w_{2}=w_{3}=w_{4}=w_{5}=\frac{1}{5}$, $\gamma^{\circ}=P(0.7,0.4)$ and $\gamma_{\circ}=P(0.25,0.85)$ are Pythagorean fuzzy numbers. By Definition 3.5, we have the Pythagorean fuzzy weighted averaging closeness index of $x_{1},x_{2},x_{3},x_{4},x_{5}$ and $x_{6}$ on $A$ as follows:
\begin{eqnarray*}
\mathscr{R}(x_{1})&=&P(\Sigma_{i=1}^{5}\mu_{\gamma_{1i}},\Sigma_{i=1}^{5}\nu_{\gamma_{1i}})=P(0.90,0.16);\\
\mathscr{R}(x_{2})&=&P(\Sigma_{i=1}^{5}\mu_{\gamma_{2i}},\Sigma_{i=1}^{5}\nu_{\gamma_{2i}})=P(0.38,0.64);\\
\mathscr{R}(x_{3})&=&P(\Sigma_{i=1}^{5}\mu_{\gamma_{3i}},\Sigma_{i=1}^{5}\nu_{\gamma_{3i}})=P(0.20,0.80);\\
\mathscr{R}(x_{4})&=&P(\Sigma_{i=1}^{5}\mu_{\gamma_{4i}},\Sigma_{i=1}^{5}\nu_{\gamma_{4i}})=P(0.30,0.70);\\
\mathscr{R}(x_{5})&=&P(\Sigma_{i=1}^{5}\mu_{\gamma_{5i}},\Sigma_{i=1}^{5}\nu_{\gamma_{5i}})=P(0.36,0.70);\\
\mathscr{R}(x_{6})&=&P(\Sigma_{i=1}^{5}\mu_{\gamma_{6i}},\Sigma_{i=1}^{5}\nu_{\gamma_{6i}})=P(0.48,0.58).
\end{eqnarray*}

By Definition 3.6, we have
\begin{eqnarray*}
POA_{(\gamma^{\circ},\gamma_{\circ})}(U)=\{x_{1}\};
CTA_{(\gamma^{\circ},\gamma_{\circ})}(U)=\{x_{2},x_{4},x_{5},x_{6}\};
NEA_{(\gamma^{\circ},\gamma_{\circ})}(U)=\emptyset.
\end{eqnarray*}
\end{example}

In Example 3.7, we classify $\{x_{1},x_{2},x_{4},x_{5},x_{6}\}$ into $POA_{(\gamma^{\circ},\gamma_{\circ})}(U),CTA_{(\gamma^{\circ},\gamma_{\circ})}(U),$ and $NEA_{(\gamma^{\circ},\gamma_{\circ})}(U)$. But we can not compare $\mathscr{R}(x_{3})=P(0.20,0.80)$ and $\gamma_{\circ}=P(0.25,0.85)$ by Definition 2.2. Therefore, we can not classify $x_{3}$ into $POA_{(\gamma^{\circ},\gamma_{\circ})}(U),CTA_{(\gamma^{\circ},\gamma_{\circ})}(U),$ and $NEA_{(\gamma^{\circ},\gamma_{\circ})}(U)$.

\begin{definition}
Let $S=(U, A, V, f)$ be a Pythagorean fuzzy information system, and $-1\leq \beta
\leq\alpha\leq 1$. Then the positive alliance, central alliance, and
negative alliance are defined as follows:
\begin{eqnarray*}
POA_{(S,\alpha,\beta)}(U)&=&\{x\in U\mid S(\mathscr{R}(x))\geq \alpha\};\\
CTA_{(S,\alpha,\beta)}(U)&=&\{x\in U\mid \beta< S(\mathscr{R}(x))<\alpha\};\\
NEA_{(S,\alpha,\beta)}(U)&=&\{x\in U\mid S(\mathscr{R}(x))\leq \beta\}.
\end{eqnarray*}
\end{definition}

By Definition 3.8, we get the positive, central, and negative alliances using two numbers $\alpha$ and $\beta$, and partition the universe into three disjoint regions, where $1\geq\alpha>\beta\geq-1$. Especially, we have $POA_{(S,\alpha,\beta)}(U)\cup CTA_{(S,\alpha,\beta)}(U)\cup NEA_{(S,\alpha,\beta)}(U)=U$.

\begin{example}(Continuation from Example 3.7) Taking $\alpha=0.5,\beta=-0.5$, by Definition 2.3, we have
\begin{eqnarray*}
S(\mathscr{R}(x_{1}))&=&0.90^{2}-0.16^{2}=0.7844;\\
S(\mathscr{R}(x_{2}))&=&0.38^{2}-0.64^{2}=-0.2652;\\
S(\mathscr{R}(x_{3}))&=&0.20^{2}-0.80^{2}=-0.6000;\\
S(\mathscr{R}(x_{4}))&=&0.30^{2}-0.70^{2}=-0.4000;\\
S(\mathscr{R}(x_{5}))&=&0.36^{2}-0.70^{2}=-0.3604;\\
S(\mathscr{R}(x_{6}))&=&0.48^{2}-0.58^{2}=-0.1060.
\end{eqnarray*}

By Definition 3.8, we have
\begin{eqnarray*}
POA_{(S,\alpha,\beta)}(U)=\{x_{1}\};
CTA_{(S,\alpha,\beta)}(U)=\{x_{2},x_{4},x_{5},x_{6}\};
NEA_{(S,\alpha,\beta)}(U)=\{x_{3}\}.
\end{eqnarray*}
\end{example}

\begin{definition}
Let $S=(U, A, V, f)$ be a Pythagorean fuzzy information system, and $0\leq \beta
\leq\alpha\leq 1$. Then the positive alliance, central alliance, and
negative alliance are defined as follows:
\begin{eqnarray*}
POA_{(\mathscr{P},\alpha,\beta)}(U)&=&\{x\in U\mid \mathscr{P}(\mathscr{R}(x))\geq \alpha\};\\
CTA_{(\mathscr{P},\alpha,\beta)}(U)&=&\{x\in U\mid \beta< \mathscr{P}(\mathscr{R}(x))<\alpha\};\\
NEA_{(\mathscr{P},\alpha,\beta)}(U)&=&\{x\in U\mid \mathscr{P}(\mathscr{R}(x))\leq \beta\}.
\end{eqnarray*}
\end{definition}

By Definition 3.10, we get the positive, central, and negative alliances using two numbers $\alpha$ and $\beta$, and partition the universe into three disjoint regions, where $1\geq\alpha>\beta\geq 0$. Especially, we have $POA_{(\mathscr{P},\alpha,\beta)}(U)\cup CTA_{(\mathscr{P},\alpha,\beta)}(U)\cup NEA_{(\mathscr{P},\alpha,\beta)}(U)=U$.

\begin{example}(Continuation from Example 3.9)
Taking $\alpha=0.75,\beta=0.3$. By Definition 2.7, we have
\begin{eqnarray*}
\mathscr{P}(\mathscr{R}(x_{1}))&=&\frac{1-0.16^{2}}{2-0.90^{2}-0.16^{2}}=0.8368;\\
\mathscr{P}(\mathscr{R}(x_{2}))&=&\frac{1-0.64^{2}}{2-0.38^{2}-0.64^{2}}=0.4083;\\
\mathscr{P}(\mathscr{R}(x_{3}))&=&\frac{1-0.80^{2}}{2-0.20^{2}-0.80^{2}}=0.2727;\\
\mathscr{P}(\mathscr{R}(x_{4}))&=&\frac{1-0.70^{2}}{2-0.70^{2}-0.30^{2}}=0.3592;\\
\mathscr{P}(\mathscr{R}(x_{5}))&=&\frac{1-0.70^{2}}{2-0.70^{2}-0.36^{2}}=0.3695;\\
\mathscr{P}(\mathscr{R}(x_{6}))&=&\frac{1-0.58^{2}}{2-0.58^{2}-0.48^{2}}=0.4584.
\end{eqnarray*}

By Definition 3.10, we have
\begin{eqnarray*}
POA_{(\mathscr{P},\alpha,\beta)}(U)=\{x_{1}\};
CTA_{(\mathscr{P},\alpha,\beta)}(U)=\{x_{2},x_{4},x_{5},x_{6}\};
NEA_{(\mathscr{P},\alpha,\beta)}(U)=\{x_{3}\}.
\end{eqnarray*}
\end{example}

By Definition 3.6, 3.8, and 3.10, we partition the universe into three regions: positive, central, and negative alliances with different types of operators, and denote the positive, central, and negative alliances of $U$ as $POA(U),CTA(U)$, and $NEA(U)$ for simplicity.

\section{Conflict Analysis based on Bayesian Minimum Risk Theory}

In this section, we investigate conflict analysis of Pythagorean fuzzy information systems based on Bayesian minimum risk theory.

\begin{definition}
A Pythagorean fuzzy loss function is a 3-tuple $\lambda=(\Omega, \mathscr{A}, \mathscr{L})$ shown as Table 2, where
$\Omega=\{X,\neg X\}$,  and $\mathscr{A}=\{a_{P},a_{B},a_{N}\}$, $X$ and $\neg X$ indicate that an object is in $X$ and not in $X$, respectively; $a_{P},a_{B}$, and $a_{N}$ denote three actions in classifying an object $x$ into $POA(U)$, $CTA(U)$, and $NEA(U)$, respectively; $\lambda_{PP},\lambda_{BP}$, and $\lambda_{NP}$ stand for the losses of taking actions $a_{P},a_{B}$, and $a_{N}$, respectively, when an object belongs to $X$; $\lambda_{PN},\lambda_{BN}$, and $\lambda_{NN}$ mean the losses of taking actions $a_{P},a_{B}$, and $a_{N}$, respectively, when an object belongs to $\neg X$, where $\lambda_{PP},\lambda_{BP}$, $\lambda_{NP}$, $\lambda_{PN},\lambda_{BN}$, and $\lambda_{NN}$ are Pythagorean fuzzy numbers.
\end{definition}

\begin{table}[H]\renewcommand{\arraystretch}{1.1}
\caption{A Pythagorean Fuzzy Loss Function.}
 \tabcolsep0.58in
\begin{tabular}{c c c }
\hline  Action & $X$ &$\neg X$\\
\hline
$a_{P}$ & $\lambda_{PP}=P(\mu_{\lambda_{PP}},\nu_{\lambda_{PP}})$& $\lambda_{PN}=P(\mu_{\lambda_{PN}},\nu_{\lambda_{PN}})$  \\
$a_{B}$ & $\lambda_{BP}=P(\mu_{\lambda_{BP}},\nu_{\lambda_{BP}})$& $\lambda_{BN}=P(\mu_{\lambda_{BN}},\nu_{\lambda_{BN}})$ \\
$a_{N}$ & $\lambda_{NP}=P(\mu_{\lambda_{NP}},\nu_{\lambda_{NP}})$& $\lambda_{NN}=P(\mu_{\lambda_{NN}},\nu_{\lambda_{NN}})$  \\
\hline
\end{tabular}
\end{table}

There are three types of Pythagorean fuzzy loss functions as follows: (1) the loss function satisfying $\lambda_{PP}\leq\lambda_{BP}\leq\lambda_{NP}$ and
$\lambda_{NN}\leq\lambda_{BN}\leq\lambda_{PN}$; (2) the loss function satisfying $S(\lambda_{PP})\leq S(\lambda_{BP})\leq S(\lambda_{NP})$ and
$S(\lambda_{NN})\leq S(\lambda_{BN})\leq S(\lambda_{PN})$; (3) the loss function satisfying $\mathscr{P}(\lambda_{PP})\leq \mathscr{P}(\lambda_{BP})\leq \mathscr{P}(\lambda_{NP})$ and
$\mathscr{P}(\lambda_{NN})\leq \mathscr{P}(\lambda_{BN})\leq \mathscr{P}(\lambda_{PN})$. For simplicity, we only discuss the loss function satisfying $\lambda_{PP}\leq\lambda_{BP}\leq\lambda_{NP}$ and
$\lambda_{NN}\leq\lambda_{BN}\leq\lambda_{PN}$ in this section.

\begin{example}
Table 3 depicts a Pythagorean fuzzy loss function, and $\lambda_{PP},\lambda_{BP},\lambda_{NP},\lambda_{NN},\lambda_{BN},$ and $\lambda_{PN}$ are Pythagorean fuzzy numbers. Especially, we have
$\lambda_{PP}\leq\lambda_{BP}\leq\lambda_{NP}$ and
$\lambda_{NN}\leq\lambda_{BN}\leq\lambda_{PN}$.
\begin{table}[H]\renewcommand{\arraystretch}{1.1}
\caption{A Pythagorean Fuzzy Loss Function.}
 \tabcolsep0.65in
\begin{tabular}{c c c }
\hline  Action & $X$ &$\neg X$\\
\hline
$a_{P}$ & $\lambda_{PP}=P(0.1,0.8)$& $\lambda_{PN}=P(0.9,0.2)$  \\
$a_{B}$ & $\lambda_{BP}=P(0.6,0.5)$& $\lambda_{BN}=P(0.5,0.6)$ \\
$a_{N}$ & $\lambda_{NP}=P(0.9,0.3)$& $\lambda_{NN}=P(0.2,0.8)$  \\
\hline
\end{tabular}
\end{table}
\end{example}

Suppose $\lambda_{PP},\lambda_{BP}$, $\lambda_{NP}$, $\lambda_{PN},\lambda_{BN}$, and $\lambda_{NN}$ are Pythagorean fuzzy numbers, which satisfy
$\lambda_{PP}\leq\lambda_{BP}\leq\lambda_{NP}$ and
$\lambda_{NN}\leq\lambda_{BN}\leq\lambda_{PN}$. For the object $x\in U$, the expected losses $R(a_{P}|x)$, $R(a_{B}|x)$, and $R(a_{N}|x)$ under the actions $a_{P},a_{B},$ and $a_{N}$, respectively, as follows:
\begin{eqnarray*}
R(a_{P}|x)&=&\mathscr{P}(\mathscr{R}(x))\ast \lambda_{PP} \oplus [1-\mathscr{P}(\mathscr{R}(x))]\ast\lambda_{PN};\\
R(a_{B}|x)&=&\mathscr{P}(\mathscr{R}(x))\ast \lambda_{BP} \oplus [1-\mathscr{P}(\mathscr{R}(x))]\ast\lambda_{BN};\\
R(a_{N}|x)&=&\mathscr{P}(\mathscr{R}(x))\ast \lambda_{NP} \oplus [1-\mathscr{P}(\mathscr{R}(x))]\ast\lambda_{NN}.
\end{eqnarray*}

According to Definitions 2.1 and 2.6, we have the expected losses $R(a_{P}|x)$, $R(a_{B}|x)$, and $R(a_{N}|x)$ as follows:
\begin{eqnarray*}
R(a_{P}|x)&=&P(\sqrt{1-(1-\mu^{2}_{\lambda_{PP}})^{\mathscr{P}(\mathscr{R}(x))}},(\nu_{\lambda_{PP}})^{\mathscr{P}(\mathscr{R}(x))})\oplus P(\sqrt{1-(1-\mu^{2}_{\lambda_{PN}})^{1-\mathscr{P}(\mathscr{R}(x))}},(\nu_{\lambda_{PN}})^{1-\mathscr{P}(\mathscr{R}(x))});\\
R(a_{B}|x)&=&P(\sqrt{1-(1-\mu^{2}_{\lambda_{BP}})^{\mathscr{P}(\mathscr{R}(x))}},(\nu_{\lambda_{BP}})^{\mathscr{P}(\mathscr{R}(x))})\oplus P(\sqrt{1-(1-\mu^{2}_{\lambda_{BN}})^{1-\mathscr{P}(\mathscr{R}(x))}},(\nu_{\lambda_{BN}})^{1-\mathscr{P}(\mathscr{R}(x))});\\
R(a_{N}|x)&=&P(\sqrt{1-(1-\mu^{2}_{\lambda_{NP}})^{\mathscr{P}(\mathscr{R}(x))}},(\nu_{\lambda_{NP}})^{\mathscr{P}(\mathscr{R}(x))})\oplus P(\sqrt{1-(1-\mu^{2}_{\lambda_{NN}})^{1-\mathscr{P}(\mathscr{R}(x))}},(\nu_{\lambda_{NN}})^{1-\mathscr{P}(\mathscr{R}(x))}).
\end{eqnarray*}

\begin{theorem} Let $R(a_{P}|x)$, $R(a_{B}|x)$, and $R(a_{N}|x)$ be the expected losses under the actions $a_{P},a_{B},$ and $a_{N}$, respectively, for the object $x\in U$. Then
\begin{eqnarray*}
R(a_{P}|x)&=&P(\sqrt{1-(1-\mu^{2}_{\lambda_{PP}})^{\mathscr{P}(\mathscr{R}(x))}\ast (1-\mu^{2}_{\lambda_{PN}})^{1-\mathscr{P}(\mathscr{R}(x))}},(\nu_{\lambda_{PP}})^{\mathscr{P}(\mathscr{R}(x))}\ast (\nu_{\lambda_{PN}})^{1-\mathscr{P}(\mathscr{R}(x))});\\
R(a_{B}|x)&=&P(\sqrt{1-(1-\mu^{2}_{\lambda_{BP}})^{\mathscr{P}(\mathscr{R}(x))}\ast (1-\mu^{2}_{\lambda_{BN}})^{1-\mathscr{P}(\mathscr{R}(x))}},(\nu_{\lambda_{BP}})^{\mathscr{P}(\mathscr{R}(x))}\ast (\nu_{\lambda_{BN}})^{1-\mathscr{P}(\mathscr{R}(x))});\\
R(a_{N}|x)&=&P(\sqrt{1-(1-\mu^{2}_{\lambda_{NP}})^{\mathscr{P}(\mathscr{R}(x))}\ast (1-\mu^{2}_{\lambda_{NN}})^{1-\mathscr{P}(\mathscr{R}(x))}},(\nu_{\lambda_{NP}})^{\mathscr{P}(\mathscr{R}(x))}\ast (\nu_{\lambda_{NN}})^{1-\mathscr{P}(\mathscr{R}(x))}).
\end{eqnarray*}
\end{theorem}

\noindent\textbf{Proof:} We assume $t_{1}=(1-\mu^{2}_{\lambda_{\bullet P}})^{\mathscr{P}(\mathscr{R}(x))},t_{2} =(1-\mu^{2}_{\lambda_{\bullet N}})^{1-\mathscr{P}(\mathscr{R}(x))},y_{1}=(\nu_{\lambda_{\bullet P}})^{\mathscr{P}(\mathscr{R}(x))}$ and $ y_{2}=(\nu_{\lambda_{\bullet N}})^{1-\mathscr{P}(\mathscr{R}(x))},$ where $\bullet= P, B, N$. By Definition 2.1, we have
\begin{eqnarray*}
R(a_{\bullet}|x)&=&P(\sqrt{1-(1-\mu^{2}_{\lambda_{\bullet P}})^{\mathscr{P}(\mathscr{R}(x))}},(\nu_{\lambda_{\bullet P}})^{\mathscr{P}(\mathscr{R}(x))})\oplus P(\sqrt{1-(1-\mu^{2}_{\lambda_{\bullet N}})^{1-\mathscr{P}(\mathscr{R}(x))}},(\nu_{\lambda_{\bullet N}})^{1-\mathscr{P}(\mathscr{R}(x))})\\
&=&P(\sqrt{1-t_{1}},y_{1})\oplus P(\sqrt{1-t_{2}},y_{2})\\
&=&P(\sqrt{1-t_{1}+1-t_{2}-(1-t_{1})(1-t_{2})},y_{1}y_{2})\\
&=&P(\sqrt{1-t_{1}t_{2}},y_{1}y_{2})\\
&=&P(\sqrt{1-(1-\mu^{2}_{\lambda_{\bullet P}})^{\mathscr{P}(\mathscr{R}(x))}\ast (1-\mu^{2}_{\lambda_{\bullet N}})^{1-\mathscr{P}(\mathscr{R}(x))}},(\nu_{\lambda_{\bullet P}})^{\mathscr{P}(\mathscr{R}(x))}\ast (\nu_{\lambda_{\bullet N}})^{1-\mathscr{P}(\mathscr{R}(x))}).\Box
\end{eqnarray*}

\begin{definition}
Let $S=(U, A, V, f)$ be a Pythagorean fuzzy information system, $R(a_{P}|x)$, $R(a_{B}|x)$, and $R(a_{N}|x)$ are the expected losses under the actions $a_{P},a_{B},$ and $a_{N}$, respectively, for the object $x\in U$. Then the Expected Loss Matrix $R(S)$ is defined as follows:
\begin{table}[H]\renewcommand{\arraystretch}{1.1}
\caption{The Expected Loss Matrix $R(S)$.}
 \tabcolsep0.55in
\begin{tabular}{c c c c }
\hline  Action & $P$ & $B$ & $N$\\
\hline
$x_{1}$ & $R(a_{P}|x_{1})$& $R(a_{B}|x_{1})$& $R(a_{N}|x_{1})$  \\
$x_{2}$ & $R(a_{P}|x_{2})$& $R(a_{B}|x_{2})$& $R(a_{N}|x_{2})$  \\
$.$ & $.$ & $.$ & $.$  \\
$.$ & $.$ & $.$ & $.$  \\
$.$ & $.$ & $.$ & $.$  \\
$x_{n}$ & $R(a_{P}|x_{n})$& $R(a_{B}|x_{n})$& $R(a_{N}|x_{n})$  \\
\hline
\end{tabular}
\end{table}
\end{definition}

\begin{theorem}
Let $S=(U, A, V, f)$ be a Pythagorean fuzzy information system, $R(a_{P}|x)$, $R(a_{B}|x)$, and $R(a_{N}|x)$ are the expected losses under the actions $a_{P},a_{B},$ and $a_{N}$, respectively, for the object $x\in U$. Then

(P1) If $R(a_{P}|x)\leq R(a_{B}|x)$  and $R(a_{P}|x)\leq R(a_{N}|x)$, then we have $x\in POA(U)$;

(B1) If $R(a_{B}|x)\leq R(a_{P}|x)$  and $R(a_{B}|x)\leq R(a_{N}|x)$, then we have $x\in CTA(U)$;

(N1) If $R(a_{N}|x)\leq R(a_{P}|x)$  and $R(a_{N}|x)\leq R(a_{B}|x)$, then we have $x\in NEA(U)$.
\end{theorem}

\noindent\textbf{Proof:} The proof is straightforward by Bayesian minimum risk theory.$\Box$

\begin{example}(Continuation from Examples 3.7 and 4.2) First, by Table 3 and Theorem 4.3, for $x_{i}\in U$, we have
\begin{eqnarray*}
R(a_{P}|x_{i})&=&P(\sqrt{1-(1-\mu^{2}_{\lambda_{PP}})^{\mathscr{P}(\mathscr{R}(x_{i}))}\ast (1-\mu^{2}_{\lambda_{PN}})^{1-\mathscr{P}(\mathscr{R}(x_{i}))}},(\nu_{\lambda_{PP}})^{\mathscr{P}(\mathscr{R}(x_{i}))}\ast (\nu_{\lambda_{PN}})^{1-\mathscr{P}(\mathscr{R}(x_{i}))})\\
&=&P(\sqrt{1-(1-0.1^{2})^{\mathscr{P}(\mathscr{R}(x_{i}))}\ast (1-0.9^{2})^{1-\mathscr{P}(\mathscr{R}(x_{i}))}},(0.8)^{\mathscr{P}(\mathscr{R}(x_{i}))}\ast (0.2)^{1-\mathscr{P}(\mathscr{R}(x_{i}))});\\
R(a_{B}|x_{i})&=&P(\sqrt{1-(1-\mu^{2}_{\lambda_{BP}})^{\mathscr{P}(\mathscr{R}(x_{i}))}\ast (1-\mu^{2}_{\lambda_{BN}})^{1-\mathscr{P}(\mathscr{R}(x_{i}))}},(\nu_{\lambda_{BP}})^{\mathscr{P}(\mathscr{R}(x_{i}))}\ast (\nu_{\lambda_{BN}})^{1-\mathscr{P}(\mathscr{R}(x_{i}))})\\
&=&P(\sqrt{1-(1-0.6^{2})^{\mathscr{P}(\mathscr{R}(x_{i}))}\ast (1-0.5^{2})^{1-\mathscr{P}(\mathscr{R}(x_{i}))}},(0.5)^{\mathscr{P}(\mathscr{R}(x_{i}))}\ast (0.6)^{1-\mathscr{P}(\mathscr{R}(x_{i}))});\\
R(a_{N}|x_{i})&=&P(\sqrt{1-(1-\mu^{2}_{\lambda_{NP}})^{\mathscr{P}(\mathscr{R}(x_{i}))}\ast (1-\mu^{2}_{\lambda_{NN}})^{1-\mathscr{P}(\mathscr{R}(x_{i}))}},(\nu_{\lambda_{NP}})^{\mathscr{P}(\mathscr{R}(x_{i}))}\ast (\nu_{\lambda_{NN}})^{1-\mathscr{P}(\mathscr{R}(x_{i}))})\\
&=&P(\sqrt{1-(1-0.9^{2})^{\mathscr{P}(\mathscr{R}(x_{i}))}\ast (1-0.2^{2})^{1-\mathscr{P}(\mathscr{R}(x_{i}))}},(0.3)^{\mathscr{P}(\mathscr{R}(x_{i}))}\ast (0.8)^{1-\mathscr{P}(\mathscr{R}(x_{i}))}).
\end{eqnarray*}

Second, by Definition 4.3, we have the Expected Loss Matrix $R(S)$ as follows:
\begin{table}[H]\renewcommand{\arraystretch}{1.1}
\caption{The Expected Loss Matrix $R(S)$.}
 \tabcolsep0.35in
\begin{tabular}{c c c c }
\hline  Action & $P$ & $B$ & $N$\\
\hline
$x_{1}$ & P(0.4937,0.6380) & P(0.5859,0.5151) & P(0.8675,0.3521)\\
$x_{2}$ & P(0.7920,0.3522) & P(0.5450,0.5570) & P(0.7103,0.5360)\\
$x_{3}$ & P(0.8378,0.2919) & P(0.5308,0.5709) & P(0.6187,0.6122)\\
$x_{4}$ & P(0.8101,0.3291) & P(0.5399,0.5620) & P(0.6808,0.5625)\\
$x_{5}$ & P(0.8065,0.3338) & P(0.5410,0.5609) & P(0.6873,0.5568)\\
$x_{6}$ & P(0.7694,0.3800) & P(0.5505,0.5514) & P(0.7393,0.5080)\\
\hline
\end{tabular}
\end{table}

Third, by Definition 4.5, we have
\begin{eqnarray*}
POA(U)&=&\{x_{1}\};
CTA(U)=\{x_{2},x_{5},x_{6}\};
NEA(U)=\emptyset.
\end{eqnarray*}
\end{example}

In Example 4.6, we can not compare the expected losses under the actions $a_{P},a_{B},$ and $a_{N}$ for the agents $x_{3}$ and $x_{4}$ by Definition 2.2. Therefore, we can not classify the agents $x_{3}$ and $x_{4}$ into $POA(U),CTA(U),$ and $NEA(U)$.

\begin{definition}
Let $S=(U, A, V, f)$ be a Pythagorean fuzzy information system, $R(a_{P}|x)$, $R(a_{B}|x)$, and $R(a_{N}|x)$ are the expected losses under the actions $a_{P},a_{B},$ and $a_{N}$, respectively, for the object $x\in U$. Then the Score Matrix $SR(S)$ is defined as follows:
\begin{table}[H]\renewcommand{\arraystretch}{1.1}
\caption{The Score Matrix $SR(S)$.}
 \tabcolsep0.48in
\begin{tabular}{c c c c }
\hline  Action & $P$ & $B$ & $N$\\
\hline
$x_{1}$ & $S(R(a_{P}|x_{1}))$& $S(R(a_{B}|x_{1}))$& $S(R(a_{N}|x_{1}))$  \\
$x_{2}$ & $S(R(a_{P}|x_{2}))$& $S(R(a_{B}|x_{2}))$& $S(R(a_{N}|x_{2}))$  \\
$.$ & $.$ & $.$ & $.$  \\
$.$ & $.$ & $.$ & $.$  \\
$.$ & $.$ & $.$ & $.$  \\
$x_{n}$ & $S(R(a_{P}|x_{n}))$& $S(R(a_{B}|x_{n}))$& $S(R(a_{N}|x_{n}))$  \\
\hline
\end{tabular}
\end{table}
\end{definition}

\begin{theorem}
Let $S=(U, A, V, f)$ be a Pythagorean fuzzy information system, $R(a_{P}|x)$, $R(a_{B}|x)$, and $R(a_{N}|x)$ are the expected losses under the actions $a_{P},a_{B},$ and $a_{N}$, respectively, for the object $x\in U$.

(P2) If $S(R(a_{P}|x))\leq S(R(a_{B}|x))$  and $S(R(a_{P}|x))\leq S(R(a_{N}|x))$, then we have $x\in POA(U)$;

(B2) If $S(R(a_{B}|x))\leq S(R(a_{P}|x))$  and $S(R(a_{B}|x))\leq S(R(a_{N}|x))$, then we have $x\in CTA(U)$;

(N2) If $S(R(a_{N}|x))\leq S(R(a_{P}|x))$  and $S(R(a_{N}|x))\leq S(R(a_{B}|x))$, then we have $x\in POA(U)$,
where
\begin{eqnarray*}
S(R(a_{P}|x))&=&1-(1-\mu^{2}_{\lambda_{PP}})^{\mathscr{P}(\mathscr{R}(x))}\ast (1-\mu^{2}_{\lambda_{PN}})^{1-\mathscr{P}(\mathscr{R}(x))}-(\nu_{\lambda_{PP}})^{2\mathscr{P}(\mathscr{R}(x))}\ast (\nu_{\lambda_{PN}})^{2-2\mathscr{P}(\mathscr{R}(x))};\\
S(R(a_{B}|x))&=&1-(1-\mu^{2}_{\lambda_{BP}})^{\mathscr{P}(\mathscr{R}(x))}\ast (1-\mu^{2}_{\lambda_{BN}})^{1-\mathscr{P}(\mathscr{R}(x))}-(\nu_{\lambda_{BP}})^{2\mathscr{P}(\mathscr{R}(x))}\ast (\nu_{\lambda_{BN}})^{2-2\mathscr{P}(\mathscr{R}(x))};\\
S(R(a_{N}|x))&=&1-(1-\mu^{2}_{\lambda_{NP}})^{\mathscr{P}(\mathscr{R}(x))}\ast (1-\mu^{2}_{\lambda_{NN}})^{1-\mathscr{P}(\mathscr{R}(x))}-(\nu_{\lambda_{NP}})^{2\mathscr{P}(\mathscr{R}(x))}\ast (\nu_{\lambda_{NN}})^{2-2\mathscr{P}(\mathscr{R}(x))}.
\end{eqnarray*}
\end{theorem}

\noindent\textbf{Proof:} The proof is straightforward by Bayesian minimum risk theory.$\Box$

\begin{example}(Continuation from Example 4.6) First, by Table 3, for $x_{i}\in U$, we have
\begin{eqnarray*}
S(R(a_{P}|x_{i}))&=&S(P(\sqrt{1-(1-\mu^{2}_{\lambda_{PP}})^{\mathscr{P}(\mathscr{R}(x_{i}))}\ast (1-\mu^{2}_{\lambda_{PN}})^{1-\mathscr{P}(\mathscr{R}(x_{i}))}},(\nu_{\lambda_{PP}})^{\mathscr{P}(\mathscr{R}(x_{i}))}\ast (\nu_{\lambda_{PN}})^{1-\mathscr{P}(\mathscr{R}(x_{i}))}))\\
&=&S(P(\sqrt{1-(1-0.1^{2})^{\mathscr{P}(\mathscr{R}(x_{i}))}\ast (1-0.9^{2})^{1-\mathscr{P}(\mathscr{R}(x_{i}))}},(0.8)^{\mathscr{P}(\mathscr{R}(x_{i}))}\ast (0.2)^{1-\mathscr{P}(\mathscr{R}(x_{i}))});\\
S(R(a_{B}|x_{i}))&=&S(P(\sqrt{1-(1-\mu^{2}_{\lambda_{BP}})^{\mathscr{P}(\mathscr{R}(x_{i}))}\ast (1-\mu^{2}_{\lambda_{BN}})^{1-\mathscr{P}(\mathscr{R}(x_{i}))}},(\nu_{\lambda_{BP}})^{\mathscr{P}(\mathscr{R}(x_{i}))}\ast (\nu_{\lambda_{BN}})^{1-\mathscr{P}(\mathscr{R}(x_{i}))}))\\
&=&S(P(\sqrt{1-(1-0.6^{2})^{\mathscr{P}(\mathscr{R}(x_{i}))}\ast (1-0.5^{2})^{1-\mathscr{P}(\mathscr{R}(x_{i}))}},(0.5)^{\mathscr{P}(\mathscr{R}(x_{i}))}\ast (0.6)^{1-\mathscr{P}(\mathscr{R}(x_{i}))}));\\
S(R(a_{N}|x_{i}))&=&S(P(\sqrt{1-(1-\mu^{2}_{\lambda_{NP}})^{\mathscr{P}(\mathscr{R}(x_{i}))}\ast (1-\mu^{2}_{\lambda_{NN}})^{1-\mathscr{P}(\mathscr{R}(x_{i}))}},(\nu_{\lambda_{NP}})^{\mathscr{P}(\mathscr{R}(x_{i}))}\ast (\nu_{\lambda_{NN}})^{1-\mathscr{P}(\mathscr{R}(x_{i}))}))\\
&=&S(P(\sqrt{1-(1-0.9^{2})^{\mathscr{P}(\mathscr{R}(x_{i}))}\ast (1-0.2^{2})^{1-\mathscr{P}(\mathscr{R}(x_{i}))}},(0.3)^{\mathscr{P}(\mathscr{R}(x_{i}))}\ast (0.8)^{1-\mathscr{P}(\mathscr{R}(x_{i}))})).
\end{eqnarray*}

Second, by Definition 4.7, we have the Score Matrix $SR(S)$ as follows:
\begin{table}[H]\renewcommand{\arraystretch}{1.1}
\caption{The Score Matrix $SR(S)$.}
 \tabcolsep0.55in
\begin{tabular}{c c c c }
\hline  Action & $P$ & $B$ & $N$\\
\hline
$x_{1}$ &-0.1633 &0.0779  &0.6286\\
$x_{2}$ &0.5031  &-0.0132 &0.2172\\
$x_{3}$ &0.6168  &-0.0442 &0.0080\\
$x_{4}$ &0.5480  &-0.0243 &0.1471\\
$x_{5}$ &0.5390  &-0.0219 &0.1623\\
$x_{6}$ &0.4476  &-0.0010 &0.2885\\
\hline
\end{tabular}
\end{table}

Third, by Theorem 4.8, we have
\begin{eqnarray*}
POA(U)&=&\{x_{1}\};
CTA(U)=\{x_{2},x_{3},x_{4},x_{5},x_{6}\};
NEA(U)=\emptyset.
\end{eqnarray*}
\end{example}

\begin{definition}
Let $S=(U, A, V, f)$ be a Pythagorean fuzzy information system, $R(a_{P}|x)$, $R(a_{B}|x)$, and $R(a_{N}|x)$ are the expected losses under the actions $a_{P},a_{B},a_{N}$, respectively, for the object $x\in U$. Then the Closeness Matrix $PR(S)$ is defined as follows:
\begin{table}[H]\renewcommand{\arraystretch}{1.1}
\caption{The Closeness Matrix $PR(S)$.}
 \tabcolsep0.48in
\begin{tabular}{c c c c }
\hline  Action & $P$ & $B$ & $N$\\
\hline
$x_{1}$ & $\mathscr{P}(R(a_{P}|x_{1}))$& $\mathscr{P}(R(a_{B}|x_{1}))$& $\mathscr{P}(R(a_{N}|x_{1}))$  \\
$x_{2}$ & $\mathscr{P}(R(a_{P}|x_{2}))$& $\mathscr{P}(R(a_{B}|x_{2}))$& $\mathscr{P}(R(a_{N}|x_{2}))$  \\
$.$ & $.$ & $.$ & $.$  \\
$.$ & $.$ & $.$ & $.$  \\
$x_{n}$ & $\mathscr{P}(R(a_{P}|x_{n}))$& $\mathscr{P}(R(a_{B}|x_{n}))$& $\mathscr{P}(R(a_{N}|x_{n}))$  \\
\hline
\end{tabular}
\end{table}
\end{definition}

\begin{theorem}
Let $S=(U, A, V, f)$ be a Pythagorean fuzzy information system, $R(a_{P}|x)$, $R(a_{B}|x)$, and $R(a_{N}|x)$ are the expected losses under the actions $a_{P},a_{B},$ and $a_{N}$, respectively, for the object $x\in U$. Then

(P3) If $\mathscr{P}(R(a_{P}|x))\leq \mathscr{P}(R(a_{B}|x))$  and $\mathscr{P}(R(a_{P}|x))\leq \mathscr{P}(R(a_{N}|x))$, then we have $x\in POA(U)$;

(B3) If $\mathscr{P}(R(a_{B}|x))\leq \mathscr{P}(R(a_{P}|x))$  and $\mathscr{P}(R(a_{B}|x))\leq \mathscr{P}(R(a_{N}|x))$, then we have  $x\in CTA(U)$;

(N3) If $\mathscr{P}(R(a_{N}|x))\leq \mathscr{P}(R(a_{P}|x))$  and $\mathscr{P}(R(a_{N}|x))\leq \mathscr{P}(R(a_{B}|x))$, then we have $x\in POA(U)$,
where
\begin{eqnarray*}
\mathscr{P}(R(a_{P}|x))&=&\frac{1-(\nu_{\lambda_{PP}})^{2\mathscr{P}(\mathscr{R}(x))}\ast (\nu_{\lambda_{PN}})^{2-2\mathscr{P}(\mathscr{R}(x))}}{2-(1-\mu^{2}_{\lambda_{PP}})^{\mathscr{P}(\mathscr{R}(x))}\ast (1-\mu^{2}_{\lambda_{PN}})^{1-\mathscr{P}(\mathscr{R}(x))}-(\nu_{\lambda_{PP}})^{2\mathscr{P}(\mathscr{R}(x))}\ast (\nu_{\lambda_{PN}})^{2-2\mathscr{P}(\mathscr{R}(x))}};\\
\mathscr{P}(R(a_{B}|x))&=&\frac{1-(\nu_{\lambda_{BP}})^{2\mathscr{P}(\mathscr{R}(x))}\ast (\nu_{\lambda_{BN}})^{2-2\mathscr{P}(\mathscr{R}(x))}}{2-(1-\mu^{2}_{\lambda_{BP}})^{\mathscr{P}(\mathscr{R}(x))}\ast (1-\mu^{2}_{\lambda_{BN}})^{1-\mathscr{P}(\mathscr{R}(x))}-(\nu_{\lambda_{BP}})^{2\mathscr{P}(\mathscr{R}(x))}\ast (\nu_{\lambda_{BN}})^{2-2\mathscr{P}(\mathscr{R}(x))}};\\
\mathscr{P}(R(a_{N}|x))&=&\frac{1-(\nu_{\lambda_{NP}})^{2\mathscr{P}(\mathscr{R}(x))}\ast (\nu_{\lambda_{NN}})^{2-2\mathscr{P}(\mathscr{R}(x))}}{2-(1-\mu^{2}_{\lambda_{NP}})^{\mathscr{P}(\mathscr{R}(x))}\ast (1-\mu^{2}_{\lambda_{NN}})^{1-\mathscr{P}(\mathscr{R}(x))}-(\nu_{\lambda_{NP}})^{2\mathscr{P}(\mathscr{R}(x))}\ast (\nu_{\lambda_{NN}})^{2-2\mathscr{P}(\mathscr{R}(x))}}.
\end{eqnarray*}
\end{theorem}

\noindent\textbf{Proof:} The proof is straightforward by Bayesian minimum risk theory.$\Box$

\begin{example}(Continuation from Example 4.9) First, for $x_{i}\in U$, we have
\begin{eqnarray*}
\mathscr{P}(R(a_{P}|x_{i}))&=&\mathscr{P}(P(\sqrt{1-(1-\mu^{2}_{\lambda_{PP}})^{\mathscr{P}(\mathscr{R}(x_{i}))}\ast (1-\mu^{2}_{\lambda_{PN}})^{1-\mathscr{P}(\mathscr{R}(x_{i}))}},(\nu_{\lambda_{PP}})^{\mathscr{P}(\mathscr{R}(x_{i}))}\ast (\nu_{\lambda_{PN}})^{1-\mathscr{P}(\mathscr{R}(x_{i}))}))\\
&=&\mathscr{P}(P(\sqrt{1-(1-0.1^{2})^{\mathscr{P}(\mathscr{R}(x_{i}))}\ast (1-0.9^{2})^{1-\mathscr{P}(\mathscr{R}(x_{i}))}},(0.8)^{\mathscr{P}(\mathscr{R}(x_{i}))}\ast (0.2)^{1-\mathscr{P}(\mathscr{R}(x_{i}))});\\
\mathscr{P}(R(a_{B}|x_{i}))&=&\mathscr{P}(P(\sqrt{1-(1-\mu^{2}_{\lambda_{BP}})^{\mathscr{P}(\mathscr{R}(x_{i}))}\ast (1-\mu^{2}_{\lambda_{BN}})^{1-\mathscr{P}(\mathscr{R}(x_{i}))}},(\nu_{\lambda_{BP}})^{\mathscr{P}(\mathscr{R}(x_{i}))}\ast (\nu_{\lambda_{BN}})^{1-\mathscr{P}(\mathscr{R}(x_{i}))}))\\
&=&\mathscr{P}(P(\sqrt{1-(1-0.6^{2})^{\mathscr{P}(\mathscr{R}(x_{i}))}\ast (1-0.5^{2})^{1-\mathscr{P}(\mathscr{R}(x_{i}))}},(0.5)^{\mathscr{P}(\mathscr{R}(x_{i}))}\ast (0.6)^{1-\mathscr{P}(\mathscr{R}(x_{i}))}));\\
\mathscr{P}(R(a_{N}|x_{i}))&=&\mathscr{P}(P(\sqrt{1-(1-\mu^{2}_{\lambda_{NP}})^{\mathscr{P}(\mathscr{R}(x_{i}))}\ast (1-\mu^{2}_{\lambda_{NN}})^{1-\mathscr{P}(\mathscr{R}(x_{i}))}},(\nu_{\lambda_{NP}})^{\mathscr{P}(\mathscr{R}(x_{i}))}\ast (\nu_{\lambda_{NN}})^{1-\mathscr{P}(\mathscr{R}(x_{i}))}))\\
&=&\mathscr{P}(P(\sqrt{1-(1-0.9^{2})^{\mathscr{P}(\mathscr{R}(x_{i}))}\ast (1-0.2^{2})^{1-\mathscr{P}(\mathscr{R}(x_{i}))}},(0.3)^{\mathscr{P}(\mathscr{R}(x_{i}))}\ast (0.8)^{1-\mathscr{P}(\mathscr{R}(x_{i}))})).
\end{eqnarray*}

Second, by Definition 4.10, we have the Closeness Matrix $PR(S)$ as follows:
\begin{table}[H]\renewcommand{\arraystretch}{1.1}
\caption{The Closeness Matrix $PR(S)$.}
 \tabcolsep0.55in
\begin{tabular}{c c c c }
\hline  Action & $P$ & $B$ & $N$\\
\hline
$x_{1}$ & 0.4395  &  0.5280  &  0.7797\\
$x_{2}$ & 0.7015  &  0.4953  &  0.5899\\
$x_{3}$ & 0.7543  &  0.4841  &  0.5032\\
$x_{4}$ & 0.7218  &  0.4913  &  0.5603\\
$x_{5}$ & 0.7176  &  0.4921  &  0.5666\\
$x_{6}$ & 0.6771  &  0.4997  &  0.6207\\
\hline
\end{tabular}
\end{table}

Third, by Theorem 4.11, we have
\begin{eqnarray*}
POA(U)&=&\{x_{1}\};
CTA(U)=\{x_{2},x_{3},x_{4},x_{5},x_{6}\};
NEA(U)=\emptyset.
\end{eqnarray*}

\end{example}

\section{Group Conflict Analysis based on Bayesian Minimum Risk Theory}

In this section, we study group conflict analysis of Pythagorean fuzzy information systems based on Bayesian minimum risk theory.

Suppose $\lambda^{(i)}_{PP},\lambda^{(i)}_{BP}$, $\lambda^{(i)}_{NP}$, $\lambda^{(i)}_{PN},\lambda^{(i)}_{BN}$, and $\lambda^{(i)}_{NN}$ are Pythagorean fuzzy numbers shown in Table 10, which satisfy
$\lambda^{(i)}_{PP}\leq\lambda^{(i)}_{BP}\leq\lambda^{(i)}_{NP}$ and
$\lambda^{(i)}_{NN}\leq\lambda^{(i)}_{BN}\leq\lambda^{(i)}_{PN}$. For the object $x\in U$, the expected losses $R^{(i)}(a_{P}|x)$, $R^{(i)}(a_{B}|x)$, and $R^{(i)}(a_{N}|x)$ under the actions $a_{P},a_{B},$ and $a_{N}$, respectively, as follows:
\begin{eqnarray*}
R^{(i)}(a_{P}|x)&=&\mathscr{P}(\mathscr{R}(x))\ast \lambda^{(i)}_{PP} \oplus [1-\mathscr{P}(\mathscr{R}(x))]\ast\lambda^{(i)}_{PN};\\
R^{(i)}(a_{B}|x)&=&\mathscr{P}(\mathscr{R}(x))\ast \lambda^{(i)}_{BP} \oplus [1-\mathscr{P}(\mathscr{R}(x))]\ast\lambda^{(i)}_{BN};\\
R^{(i)}(a_{N}|x)&=&\mathscr{P}(\mathscr{R}(x))\ast \lambda^{(i)}_{NP} \oplus [1-\mathscr{P}(\mathscr{R}(x))]\ast\lambda^{(i)}_{NN}.
\end{eqnarray*}
\begin{table}[H]\renewcommand{\arraystretch}{1.2}
\caption{Pythagorean Fuzzy Loss Functions $\{\lambda^{(i)}|i=1,2,...,m\}$.} \tabcolsep0.4in
\label{bigtable}
\begin{center}
\begin{tabular}{ c c c c c}
\hline
$\lambda$ & Action & $X$ &$\neg X$\\
\hline
\multirow{3}*{$\lambda^{(1)}$} & $a_{P}$ & $\lambda^{(1)}_{PP}=P(\mu_{\lambda^{(1)}_{PP}},\nu_{\lambda^{(1)}_{PP}})$& $\lambda^{(1)}_{PN}=P(\mu_{\lambda^{(1)}_{PN}},\nu_{\lambda^{(1)}_{PN}})$  \\
&$a_{B}$ & $\lambda^{(1)}_{BP}=P(\mu_{\lambda^{(1)}_{BP}},\nu_{\lambda^{(1)}_{BP}})$& $\lambda^{(1)}_{BN}=P(\mu_{\lambda^{(1)}_{BN}},\nu_{\lambda^{(1)}_{BN}})$ \\
&$a_{N}$ & $\lambda^{(1)}_{NP}=P(\mu_{\lambda^{(1)}_{NP}},\nu_{\lambda^{(1)}_{NP}})$& $\lambda^{(1)}_{NN}=P(\mu_{\lambda^{(1)}_{NN}},\nu_{\lambda^{(1)}_{NN}})$  \\
\hline
\multirow{3}*{$\lambda^{(2)}$} & $a_{P}$ & $\lambda^{(2)}_{PP}=P(\mu_{\lambda^{(2)}_{PP}},\nu_{\lambda^{(2)}_{PP}})$& $\lambda^{(2)}_{PN}=P(\mu_{\lambda^{(2)}_{PN}},\nu_{\lambda^{(2)}_{PN}})$  \\
&$a_{B}$ & $\lambda^{(2)}_{BP}=P(\mu_{\lambda^{(2)}_{BP}},\nu_{\lambda^{(2)}_{BP}})$& $\lambda^{(2)}_{BN}=P(\mu_{\lambda^{(2)}_{BN}},\nu_{\lambda^{(2)}_{BN}})$ \\
&$a_{N}$ & $\lambda^{(2)}_{NP}=P(\mu_{\lambda^{(2)}_{NP}},\nu_{\lambda^{(2)}_{NP}})$& $\lambda^{(2)}_{NN}=P(\mu_{\lambda^{(2)}_{NN}},\nu_{\lambda^{(2)}_{NN}})$  \\
\hline
\multirow{3}*{$.$} & $.$ & $.$& $.$  \\
&$.$ & $.$& $.$ \\
&$.$ & $.$& $.$ \\
\hline
\multirow{3}*{$.$} & $.$ & $.$& $.$  \\
&$.$ & $.$& $.$ \\
&$.$ & $.$& $.$ \\
\hline
\multirow{3}*{$.$} & $.$ & $.$& $.$  \\
&$.$ & $.$& $.$ \\
&$.$ & $.$& $.$ \\
\hline
\multirow{3}*{$\lambda^{(m)}$} & $a_{P}$ & $\lambda^{(m)}_{PP}=P(\mu_{\lambda^{(m)}_{PP}},\nu_{\lambda^{(m)}_{PP}})$& $\lambda^{(m)}_{PN}=P(\mu_{\lambda^{(m)}_{PN}},\nu_{\lambda^{(m)}_{PN}})$  \\
&$a_{B}$ & $\lambda^{(m)}_{BP}=P(\mu_{\lambda^{(m)}_{BP}},\nu_{\lambda^{(m)}_{BP}})$& $\lambda^{(m)}_{BN}=P(\mu_{\lambda^{(m)}_{BN}},\nu_{\lambda^{(m)}_{BN}})$ \\
&$a_{N}$ & $\lambda^{(m)}_{NP}=P(\mu_{\lambda^{(m)}_{NP}},\nu_{\lambda^{(m)}_{NP}})$& $\lambda^{(m)}_{NN}=P(\mu_{\lambda^{(m)}_{NN}},\nu_{\lambda^{(m)}_{NN}})$  \\
\hline
\end{tabular}
\end{center}
\end{table}

\begin{example}(Continuation from Example 4.6) Table 11 depicts a collection of Pythagorean fuzzy loss functions $\{\lambda^{(i)}|i=1,2,3\}$, which are given by a group of experts.
\begin{table}[htbp]\renewcommand{\arraystretch}{1.2}
\caption{Pythagorean Fuzzy Loss Functions $\{\lambda^{(i)}|i=1,2,3\}$.} \tabcolsep0.4in
\label{bigtable}
\begin{center}
\begin{tabular}{ c c c c c}
\hline
$\lambda$ & Action & $X$ &$\neg X$\\
\hline
\multirow{3}*{$\lambda^{(1)}$} & $a_{P}$ & $\lambda_{PP}=P(0.1,0.8)$& $\lambda_{PN}=P(0.9,0.2)$  \\
&$a_{B}$ & $\lambda_{BP}=P(0.6,0.5)$& $\lambda_{BN}=P(0.5,0.6)$ \\
&$a_{N}$ & $\lambda_{NP}=P(0.9,0.3)$& $\lambda_{NN}=P(0.2,0.8)$  \\
\hline
\multirow{3}*{$\lambda^{(2)}$} & $a_{P}$ & $\lambda_{PP}=P(0.2,0.9)$& $\lambda_{PN}=P(0.8,0.3)$  \\
&$a_{B}$ & $\lambda_{BP}=P(0.5,0.7)$& $\lambda_{BN}=P(0.6,0.5)$ \\
&$a_{N}$ & $\lambda_{NP}=P(0.8,0.2)$& $\lambda_{NN}=P(0.1,0.9)$  \\
\hline
\multirow{3}*{$\lambda^{(3)}$} & $a_{P}$ & $\lambda_{PP}=P(0.3,0.9)$& $\lambda_{PN}=P(0.8,0.1)$  \\
&$a_{B}$ & $\lambda_{BP}=P(0.5,0.6)$& $\lambda_{BN}=P(0.6,0.6)$ \\
&$a_{N}$ & $\lambda_{NP}=P(0.7,0.1)$& $\lambda_{NN}=P(0.2,0.9)$  \\
\hline
\end{tabular}
\end{center}
\end{table}

\end{example}

According to Definitions 2.1 and 2.3, we have the expected losses $R^{(i)}(a_{P}|x), R^{(i)}(a_{B}|x)$ and $R^{(i)}(a_{N}|x)$ with respect to the Pythagorean fuzzy loss function $\lambda^{(i)}$ as follows:
\begin{eqnarray*}
R^{(i)}(a_{P}|x)&=&P(\sqrt{1-(1-\mu^{2}_{\lambda^{(i)}_{PP}})^{\mathscr{P}(\mathscr{R}(x))}},(\nu_{\lambda^{(i)}_{PP}})^{\mathscr{P}(\mathscr{R}(x))})\oplus P(\sqrt{1-(1-\mu^{2}_{\lambda^{(i)}_{PN}})^{1-\mathscr{P}(\mathscr{R}(x))}},(\nu_{\lambda^{(i)}_{PN}})^{1-\mathscr{P}(\mathscr{R}(x))})\\
&=&P(\sqrt{1-(1-\mu^{2}_{\lambda^{(i)}_{PP}})^{\mathscr{P}(\mathscr{R}(x))}\ast (1-\mu^{2}_{\lambda^{(i)}_{PN}})^{1-\mathscr{P}(\mathscr{R}(x))}},(\nu_{\lambda^{(i)}_{PP}})^{\mathscr{P}(\mathscr{R}(x))}\ast (\nu_{\lambda^{(i)}_{PN}})^{1-\mathscr{P}(\mathscr{R}(x))});\\
R^{(i)}(a_{B}|x)&=&P(\sqrt{1-(1-\mu^{2}_{\lambda^{(i)}_{BP}})^{\mathscr{P}(\mathscr{R}(x))}},(\nu_{\lambda^{(i)}_{BP}})^{\mathscr{P}(\mathscr{R}(x))})\oplus P(\sqrt{1-(1-\mu^{2}_{\lambda^{(i)}_{BN}})^{1-\mathscr{P}(\mathscr{R}(x))}},(\nu_{\lambda_{BN}})^{1-\mathscr{P}(\mathscr{R}(x))})\\
&=&P(\sqrt{1-(1-\mu^{2}_{\lambda^{(i)}_{BP}})^{\mathscr{P}(\mathscr{R}(x))}\ast (1-\mu^{2}_{\lambda^{(i)}_{BN}})^{1-\mathscr{P}(\mathscr{R}(x))}},(\nu_{\lambda^{(i)}_{BP}})^{\mathscr{P}(\mathscr{R}(x))}\ast (\nu_{\lambda_{BN}})^{1-\mathscr{P}(\mathscr{R}(x))});\\
R^{(i)}(a_{N}|x)&=&P(\sqrt{1-(1-\mu^{2}_{\lambda^{(i)}_{NP}})^{\mathscr{P}(\mathscr{R}(x))}\ast (1-\mu^{2}_{\lambda^{(i)}_{NN}})^{1-\mathscr{P}(\mathscr{R}(x))}},(\nu_{\lambda^{(i)}_{NP}})^{\mathscr{P}(\mathscr{R}(x))}\ast (\nu_{\lambda^{(i)}_{NN}})^{1-\mathscr{P}(\mathscr{R}(x))})\\
&=&P(\sqrt{1-(1-\mu^{2}_{\lambda^{(i)}_{NP}})^{\mathscr{P}(\mathscr{R}(x))}\ast (1-\mu^{2}_{\lambda^{(i)}_{NN}})^{1-\mathscr{P}(\mathscr{R}(x))}},(\nu_{\lambda^{(i)}_{NP}})^{\mathscr{P}(\mathscr{R}(x))}\ast (\nu_{\lambda^{(i)}_{NN}})^{1-\mathscr{P}(\mathscr{R}(x))}).
\end{eqnarray*}

\begin{theorem}
Let $R^{(i)}(a_{P}|x)$, $R^{(i)}(a_{B}|x)$, and $R^{(i)}(a_{N}|x)$ be the expected losses under the actions $a_{P},a_{B},$ and $a_{N}$ using the Pythagorean fuzzy loss function $\lambda^{(i)}$, respectively, for the object $x\in U$, and $\mathscr{K}=\{k_{1},k_{2},...,k_{m}\}$ be the weight vector of $R^{(i)}(a_{\bullet}|x)(i=1,2,...,m, \bullet=P,B,N)$. Then
\begin{eqnarray*}
&&\mathscr{R}(R^{(1)}(a_{P}|x),...,R^{(m)}(a_{P}|x))\\&&=P(\Sigma^{m}_{i=1}k_{i}\ast\sqrt{1-(1-\mu^{2}_{\lambda^{(i)}_{PP}})^{\mathscr{P}(\mathscr{R}(x))}\ast (1-\mu^{2}_{\lambda^{(i)}_{PN}})^{1-\mathscr{P}(\mathscr{R}(x))}},\Sigma^{m}_{i=1}k_{i}\ast(\nu_{\lambda^{(i)}_{PP}})^{\mathscr{P}(\mathscr{R}(x))}\ast (\nu_{\lambda^{i}_{PN}})^{1-\mathscr{P}(\mathscr{R}(x))});\\
&&\mathscr{R}(R^{(1)}(a_{B}|x),...,R^{(m)}(a_{B}|x))\\&&=P(\Sigma^{m}_{i=1}k_{i}\ast\sqrt{1-(1-\mu^{2}_{\lambda^{(i)}_{BP}})^{\mathscr{P}(\mathscr{R}(x))}\ast (1-\mu^{2}_{\lambda^{(i)}_{BN}})^{1-\mathscr{P}(\mathscr{R}(x))}},\Sigma^{m}_{i=1}k_{i}\ast(\nu_{\lambda^{(i)}_{BP}})^{\mathscr{P}(\mathscr{R}(x))}\ast (\nu_{\lambda_{BN}})^{1-\mathscr{P}(\mathscr{R}(x))});\\
&&\mathscr{R}(R^{(1)}(a_{N}|x),...,R^{(m)}(a_{N}|x))\\&&=P(\Sigma^{m}_{i=1}k_{i}\ast\sqrt{1-(1-\mu^{2}_{\lambda^{(i)}_{NP}})^{\mathscr{P}(\mathscr{R}(x))}\ast (1-\mu^{2}_{\lambda^{(i)}_{NN}})^{1-\mathscr{P}(\mathscr{R}(x))}},\Sigma^{m}_{i=1}k_{i}\ast(\nu_{\lambda^{(i)}_{NP}})^{\mathscr{P}(\mathscr{R}(x))}\ast (\nu_{\lambda^{(i)}_{NN}})^{1-\mathscr{P}(\mathscr{R}(x))}).
\end{eqnarray*}
\end{theorem}

\noindent\textbf{Proof:} The proof is straightforward by Theorem 4.3.$\Box$

\begin{definition}
Let $S=(U, A, V, f)$ be a Pythagorean fuzzy information system, $\mathscr{R}(R^{(1)}(a_{P}|x),...,R^{(m)}(a_{P}|x))$, $\mathscr{R}(R^{(1)}(a_{B}|x),...,R^{(m)}(a_{B}|x))$, and $\mathscr{R}(R^{(1)}(a_{N}|x),...,R^{(m)}(a_{N}|x))$ are the expected losses under the actions $a_{P},a_{B},$ and $a_{N}$, respectively, for the object $x\in U$. Then the Group Expected Loss Matrix $\mathscr{R}(R(S))$ is defined as follows:
\begin{table}[H]\renewcommand{\arraystretch}{1.1}
\caption{The Group Expected Loss Matrix $\mathscr{R}(R(S))$.}
 \tabcolsep0.06in
\begin{tabular}{c c c c }
\hline  Action & $P$ & $B$ & $N$\\
\hline
$x_{1}$ & $\mathscr{R}(R^{(1)}(a_{P}|x_{1}),...,R^{(m)}(a_{P}|x_{1}))$& $\mathscr{R}(R^{(1)}(a_{B}|x_{1}),...,R^{(m)}(a_{B}|x_{1}))$& $\mathscr{R}(R^{(1)}(a_{N}|x_{1}),...,R^{(m)}(a_{N}|x_{1}))$\\
$x_{2}$ & $\mathscr{R}(R^{(1)}(a_{P}|x_{2}),...,R^{(m)}(a_{P}|x_{2}))$& $\mathscr{R}(R^{(1)}(a_{B}|x_{2}),...,R^{(m)}(a_{B}|x_{2}))$& $\mathscr{R}(R^{(1)}(a_{N}|x_{2}),...,R^{(m)}(a_{N}|x_{2}))$\\
$.$ & $.$ & $.$ & $.$  \\
$.$ & $.$ & $.$ & $.$  \\
$.$ & $.$ & $.$ & $.$  \\
$x_{n}$ & $\mathscr{R}(R^{(1)}(a_{P}|x_{n}),...,R^{(m)}(a_{P}|x_{n}))$& $\mathscr{R}(R^{(1)}(a_{B}|x_{n}),...,R^{(m)}(a_{B}|x_{n}))$& $\mathscr{R}(R^{(1)}(a_{N}|x_{n}),...,R^{(m)}(a_{N}|x_{n}))$\\
\hline
\end{tabular}
\end{table}
\end{definition}

\begin{theorem}
Let $S=(U, A, V, f)$ be a Pythagorean fuzzy information system, $R(a_{P}|x)$, $R(a_{B}|x)$, and $R(a_{N}|x)$ are the expected losses under the actions $a_{P},a_{B},a_{N}$, respectively, for the object $x\in U$. Then

(P1) If $\mathscr{R}(R^{(1)}(a_{P}|x),...,R^{(m)}(a_{P}|x))\leq \mathscr{R}(R^{(1)}(a_{B}|x),...,R^{(m)}(a_{B}|x))$  and $\mathscr{R}(R^{(1)}(a_{P}|x),...,R^{(m)}(a_{P}|x))\leq \mathscr{R}(R^{(1)}(a_{N}|x),...,R^{(m)}(a_{N}|x))$, then we have $x\in POA(U)$;

(B1) If $\mathscr{R}(R^{(1)}(a_{B}|x),...,R^{(m)}(a_{B}|x))\leq \mathscr{R}(R^{(1)}(a_{P}|x),...,R^{(m)}(a_{P}|x))$  and $\mathscr{R}(R^{(1)}(a_{B}|x),...,R^{(m)}(a_{B}|x))\leq \mathscr{R}(R^{(1)}(a_{N}|x),...,R^{(m)}(a_{N}|x))$, then we have $x\in CTA(U)$;

(N1) If $\mathscr{R}(R^{(1)}(a_{N}|x),...,R^{(m)}(a_{N}|x))\leq \mathscr{R}(R^{(1)}(a_{P}|x),...,R^{(m)}(a_{P}|x))$  and $\mathscr{R}(R^{(1)}(a_{N}|x),...,R^{(m)}(a_{N}|x))\leq \mathscr{R}(R^{(1)}(a_{B}|x),...,R^{(m)}(a_{B}|x))$, then we have $x\in NEA(U)$.
\end{theorem}

\noindent\textbf{Proof:} The proof is straightforward by Bayesian minimum risk theory.$\Box$

\begin{example}(Continuation from Example 5.1) Taking the weight vector $k_{1}=k_{2}=k_{3}=\frac{1}{3}$ for Pythagorean fuzzy loss functions $\{\lambda^{(i)}|i=1,2,3\}$. First, for $x_{j}\in U$, by Theorem 5.2, we have
\begin{eqnarray*}
&&\mathscr{R}(R^{(1)}(a_{P}|x_{j}),...,R^{(m)}(a_{P}|x_{j}))\\&&=P(\Sigma^{m}_{i=1}k_{i}\ast\sqrt{1-(1-\mu^{2}_{\lambda^{(i)}_{PP}})^{\mathscr{P}(\mathscr{R}(x_{j}))}\ast (1-\mu^{2}_{\lambda^{(i)}_{PN}})^{1-\mathscr{P}(\mathscr{R}(x_{j}))}},\Sigma^{m}_{i=1}k_{i}\ast(\nu_{\lambda^{(i)}_{PP}})^{\mathscr{P}(\mathscr{R}(x_{j}))}\ast (\nu_{\lambda^{i}_{PN}})^{1-\mathscr{P}(\mathscr{R}(x_{j}))});\\
&&\mathscr{R}(R^{(1)}(a_{B}|x_{j}),...,R^{(m)}(a_{B}|x_{j}))\\&&=P(\Sigma^{m}_{i=1}k_{i}\ast\sqrt{1-(1-\mu^{2}_{\lambda^{(i)}_{BP}})^{\mathscr{P}(\mathscr{R}(x_{j}))}\ast (1-\mu^{2}_{\lambda^{(i)}_{BN}})^{1-\mathscr{P}(\mathscr{R}(x_{j}))}},\Sigma^{m}_{i=1}k_{i}\ast(\nu_{\lambda^{(i)}_{BP}})^{\mathscr{P}(\mathscr{R}(x_{j}))}\ast (\nu_{\lambda_{BN}})^{1-\mathscr{P}(\mathscr{R}(x_{j}))});\\
&&\mathscr{R}(R^{(1)}(a_{N}|x_{j}),...,R^{(m)}(a_{N}|x_{j}))\\&&=P(\Sigma^{m}_{i=1}k_{i}\ast\sqrt{1-(1-\mu^{2}_{\lambda^{(i)}_{NP}})^{\mathscr{P}(\mathscr{R}(x_{j}))}\ast (1-\mu^{2}_{\lambda^{(i)}_{NN}})^{1-\mathscr{P}(\mathscr{R}(x_{j}))}},\Sigma^{m}_{i=1}k_{i}\ast(\nu_{\lambda^{(i)}_{NP}})^{\mathscr{P}(\mathscr{R}(x_{j}))}\ast (\nu_{\lambda^{(i)}_{NN}})^{1-\mathscr{P}(\mathscr{R}(x_{j}))}).
\end{eqnarray*}

Second, by Definition 5.3, we have the Group Expected Loss Matrix $\mathscr{R}(R(S))$ as follows:
\begin{table}[H]\renewcommand{\arraystretch}{1.1}
\caption{The Group Expected Loss Matrix $\mathscr{R}(R(S))$.}
 \tabcolsep0.4in
\begin{tabular}{c c c c }
\hline  Action & $P$ & $B$ & $N$\\
\hline
$x_{1}$ & P(0.4623,0.6731) & P(0.5412,0.5926) & P(0.7617,0.2503)\\
$x_{2}$ & P(0.7203,0.3558) & P(0.5571,0.5769) & P(0.6020,0.4633)\\
$x_{3}$ & P(0.7660,0.2929) & P(0.5609,0.5730) & P(0.5186,0.5679)\\
$x_{4}$ & P(0.7380,0.3314) & P(0.5586,0.5754) & P(0.5746,0.4986)\\
$x_{5}$ & P(0.7344,0.3364) & P(0.5583,0.5757) & P(0.5806,0.4909)\\
$x_{6}$ & P(0.6987,0.3852) & P(0.5554,0.5786) & P(0.6295,0.4273)\\
\hline
\end{tabular}
\end{table}

Third, by Theorem 5.4, we have
\begin{eqnarray*}
POA(U)&=&\{x_{1}\};
CTA(U)=\{x_{2},x_{4},x_{5},x_{6}\};
NEA(U)=\emptyset.
\end{eqnarray*}
\end{example}

In Example 5.5, we can not compare the expected losses under the actions $a_{P},a_{B},$ and $a_{N}$ for the agents $x_{3}$ by Definition 2.2. Therefore, we can not classify the agents $x_{3}$ into $POA(U),CTA(U),$ and $NEA(U)$.

\begin{definition}
Let $S=(U, A, V, f)$ be a Pythagorean fuzzy information system, $\mathscr{R}(R^{(1)}(a_{P}|x),...,R^{(m)}(a_{P}|x))$, $\mathscr{R}(R^{(1)}(a_{B}|x),...,R^{(m)}(a_{B}|x))$, and $\mathscr{R}(R^{(1)}(a_{N}|x),...,R^{(m)}(a_{N}|x))$ are the expected losses under the actions $a_{P},a_{B},a_{N}$, respectively, for the object $x\in U$. Then the Group Score Matrix $\mathscr{S}(R(S))$ is defined as follows:
\begin{table}[H]\renewcommand{\arraystretch}{1.1}
\caption{The Group Score Matrix $\mathscr{S}(R(S))$.}
 \tabcolsep0.06in
\begin{tabular}{c c c c }
\hline  Action & $P$ & $B$ & $N$\\
\hline
$x_{1}$ & $S(\mathscr{R}(R^{(1)}(a_{P}|x_{1}),...,R^{(m)}(a_{P}|x_{1})))$& $S(\mathscr{R}(R^{(1)}(a_{B}|x_{1}),...,R^{(m)}(a_{B}|x_{1})))$& $S(\mathscr{R}(R^{(1)}(a_{N}|x_{1}),...,R^{(m)}(a_{N}|x_{1})))$\\
$x_{2}$ & $S(\mathscr{R}(R^{(1)}(a_{P}|x_{2}),...,R^{(m)}(a_{P}|x_{2})))$& $S(\mathscr{R}(R^{(1)}(a_{B}|x_{2}),...,R^{(m)}(a_{B}|x_{2})))$& $S(\mathscr{R}(R^{(1)}(a_{N}|x_{2}),...,R^{(m)}(a_{N}|x_{2})))$\\
$.$ & $.$ & $.$ & $.$  \\
$.$ & $.$ & $.$ & $.$  \\
$.$ & $.$ & $.$ & $.$  \\
$x_{n}$ & $S(\mathscr{R}(R^{(1)}(a_{P}|x_{n}),...,R^{(m)}(a_{P}|x_{n})))$& $S(\mathscr{R}(R^{(1)}(a_{B}|x_{n}),...,R^{(m)}(a_{B}|x_{n})))$& $S(\mathscr{R}(R^{(1)}(a_{N}|x_{n}),...,R^{(m)}(a_{N}|x_{n})))$\\
\hline
\end{tabular}
\end{table}
\end{definition}

\begin{theorem}
Let $S=(U, A, V, f)$ be a Pythagorean fuzzy information system, $R(a_{P}|x)$, $R(a_{B}|x)$, and $R(a_{N}|x)$ are the expected losses under the actions $a_{P},a_{B},$ and $a_{N}$, respectively, for the object $x\in U$. Then

(P1) If $S(\mathscr{R}(R^{(1)}(a_{P}|x),...,R^{(m)}(a_{P}|x)))\leq S(\mathscr{R}(R^{(1)}(a_{B}|x),...,R^{(m)}(a_{B}|x)))$  and $S(\mathscr{R}(R^{(1)}(a_{P}|x),...,R^{(m)}(a_{P}|x)))\leq S(\mathscr{R}(R^{(1)}(a_{N}|x),...,R^{(m)}(a_{N}|x)))$, then we have decide $x\in POA(U)$;

(B1) If $S(\mathscr{R}(R^{(1)}(a_{B}|x),...,R^{(m)}(a_{B}|x)))\leq S(\mathscr{R}(R^{(1)}(a_{P}|x),...,R^{(m)}(a_{P}|x)))$  and $S(\mathscr{R}(R^{(1)}(a_{B}|x),...,R^{(m)}(a_{B}|x)))\leq S(\mathscr{R}(R^{(1)}(a_{N}|x),...,R^{(m)}(a_{N}|x)))$, then we have $x\in CTA(U)$;

(N1) If $S(\mathscr{R}(R^{(1)}(a_{N}|x),...,R^{(m)}(a_{N}|x)))\leq S(\mathscr{R}(R^{(1)}(a_{P}|x),...,R^{(m)}(a_{P}|x)))$  and $S(\mathscr{R}(R^{(1)}(a_{N}|x),...,R^{(m)}(a_{N}|x)))\leq S(\mathscr{R}(R^{(1)}(a_{B}|x),...,R^{(m)}(a_{B}|x)))$, then we have $x\in NEA(U)$.
\end{theorem}

\noindent\textbf{Proof:} The proof is straightforward by Bayesian minimum risk theory.$\Box$

\begin{example}(Continuation from Example 5.5) First, for $x_{j}\in U$, by Theorem 5.2, we have
\begin{eqnarray*}
&&S(\mathscr{R}(R^{(1)}(a_{P}|x_{j}),...,R^{(m)}(a_{P}|x_{j})))\\&&=S(P(\Sigma^{m}_{i=1}k_{i}\ast\sqrt{1-(1-\mu^{2}_{\lambda^{(i)}_{PP}})^{\mathscr{P}(\mathscr{R}(x_{j}))}\ast (1-\mu^{2}_{\lambda^{(i)}_{PN}})^{1-\mathscr{P}(\mathscr{R}(x_{j}))}},\Sigma^{m}_{i=1}k_{i}\ast(\nu_{\lambda^{(i)}_{PP}})^{\mathscr{P}(\mathscr{R}(x_{j}))}\ast (\nu_{\lambda^{i}_{PN}})^{1-\mathscr{P}(\mathscr{R}(x_{j}))}));\\
&&S(\mathscr{R}(R^{(1)}(a_{B}|x_{j}),...,R^{(m)}(a_{B}|x_{j})))\\&&=S(P(\Sigma^{m}_{i=1}k_{i}\ast\sqrt{1-(1-\mu^{2}_{\lambda^{(i)}_{BP}})^{\mathscr{P}(\mathscr{R}(x_{j}))}\ast (1-\mu^{2}_{\lambda^{(i)}_{BN}})^{1-\mathscr{P}(\mathscr{R}(x_{j}))}},\Sigma^{m}_{i=1}k_{i}\ast(\nu_{\lambda^{(i)}_{BP}})^{\mathscr{P}(\mathscr{R}(x_{j}))}\ast (\nu_{\lambda_{BN}})^{1-\mathscr{P}(\mathscr{R}(x_{j}))}));\\
&&S(\mathscr{R}(R^{(1)}(a_{N}|x_{j}),...,R^{(m)}(a_{N}|x_{j})))\\&&=S(P(\Sigma^{m}_{i=1}k_{i}\ast\sqrt{1-(1-\mu^{2}_{\lambda^{(i)}_{NP}})^{\mathscr{P}(\mathscr{R}(x_{j}))}\ast (1-\mu^{2}_{\lambda^{(i)}_{NN}})^{1-\mathscr{P}(\mathscr{R}(x_{j}))}},\Sigma^{m}_{i=1}k_{i}\ast(\nu_{\lambda^{(i)}_{NP}})^{\mathscr{P}(\mathscr{R}(x_{j}))}\ast (\nu_{\lambda^{(i)}_{NN}})^{1-\mathscr{P}(\mathscr{R}(x_{j}))})).
\end{eqnarray*}

Second, by Definition 5.6, we have the Group Score Matrix $\mathscr{S}(R(S))$ as follows:
\begin{table}[H]\renewcommand{\arraystretch}{1.1}
\caption{The Group Score Matrix $\mathscr{S}(R(S))$.}
 \tabcolsep0.6in
\begin{tabular}{c c c c }
\hline  Action & $P$ & $B$ & $N$\\
\hline
$x_{1}$ &-0.2393 & -0.0583  &  0.5176\\
$x_{2}$ &0.3922  & -0.0224  &  0.1478\\
$x_{3}$ &0.5009  & -0.0137  & -0.0536\\
$x_{4}$ &0.4349  & -0.0191  &  0.0816\\
$x_{5}$ &0.4263  & -0.0198  &  0.0961\\
$x_{6}$ &0.3398  & -0.0262  &  0.2137\\
\hline
\end{tabular}
\end{table}

Third, by Theorem 5.7, we have
\begin{eqnarray*}
POA(U)&=&\{x_{1}\};
CTA(U)=\{x_{2},x_{4},x_{5},x_{6}\};
NEA(U)=\{x_{3}\}.
\end{eqnarray*}
\end{example}

\begin{definition}
Let $S=(U, A, V, f)$ be a Pythagorean fuzzy information system, $\mathscr{R}(R^{(1)}(a_{P}|x),...,R^{(m)}(a_{P}|x))$, $\mathscr{R}(R^{(1)}(a_{B}|x),...,R^{(m)}(a_{B}|x))$, and $\mathscr{R}(R^{(1)}(a_{N}|x),...,R^{(m)}(a_{N}|x))$ are the expected losses under the actions $a_{P},a_{B},$ and $a_{N}$, respectively, for the object $x\in U$. Then the Group Closeness Matrix $\mathscr{P}(R(S))$ is defined as follows:
\begin{table}[H]\renewcommand{\arraystretch}{1.1}
\caption{The Group Closeness Matrix $\mathscr{P}(R(S))$.}
 \tabcolsep0.06in
\begin{tabular}{c c c c }
\hline  Action & $P$ & $B$ & $N$\\
\hline
$x_{1}$ & $\mathscr{P}(\mathscr{R}(R^{(1)}(a_{P}|x_{1}),...,R^{(m)}(a_{P}|x_{1})))$& $\mathscr{P}(\mathscr{R}(R^{(1)}(a_{B}|x_{1}),...,R^{(m)}(a_{B}|x_{1})))$& $\mathscr{P}(\mathscr{R}(R^{(1)}(a_{N}|x_{1}),...,R^{(m)}(a_{N}|x_{1})))$\\
$x_{2}$ & $\mathscr{P}(\mathscr{R}(R^{(1)}(a_{P}|x_{2}),...,R^{(m)}(a_{P}|x_{2})))$& $\mathscr{P}(\mathscr{R}(R^{(1)}(a_{B}|x_{2}),...,R^{(m)}(a_{B}|x_{2})))$& $\mathscr{P}(\mathscr{R}(R^{(1)}(a_{N}|x_{2}),...,R^{(m)}(a_{N}|x_{2})))$\\
$.$ & $.$ & $.$ & $.$  \\
$.$ & $.$ & $.$ & $.$  \\
$.$ & $.$ & $.$ & $.$  \\
$x_{n}$ & $\mathscr{P}(\mathscr{R}(R^{(1)}(a_{P}|x_{n}),...,R^{(m)}(a_{P}|x_{n})))$& $\mathscr{P}(\mathscr{R}(R^{(1)}(a_{B}|x_{n}),...,R^{(m)}(a_{B}|x_{n})))$& $\mathscr{P}(\mathscr{R}(R^{(1)}(a_{N}|x_{n}),...,R^{(m)}(a_{N}|x_{n})))$\\
\hline
\end{tabular}
\end{table}
\end{definition}

\begin{theorem}
Let $S=(U, A, V, f)$ be a Pythagorean fuzzy information system, $R(a_{P}|x)$, $R(a_{B}|x)$, and $R(a_{N}|x)$ are the expected losses under the actions $a_{P},a_{B},$ and $a_{N}$, respectively, for the object $x\in U$. Then

(P1) If $\mathscr{P}(\mathscr{R}(R^{(1)}(a_{P}|x),...,R^{(m)}(a_{P}|x)))\leq \mathscr{P}(\mathscr{R}(R^{(1)}(a_{B}|x),...,R^{(m)}(a_{B}|x)))$  and $\mathscr{P}(\mathscr{R}(R^{(1)}(a_{P}|x),...,R^{(m)}(a_{P}|x)))\leq \mathscr{P}(\mathscr{R}(R^{(1)}(a_{N}|x),...,R^{(m)}(a_{N}|x)))$, decide $x\in POA(U)$;

(B1) If $\mathscr{P}(\mathscr{R}(R^{(1)}(a_{B}|x),...,R^{(m)}(a_{B}|x)))\leq \mathscr{P}(\mathscr{R}(R^{(1)}(a_{P}|x),...,R^{(m)}(a_{P}|x)))$  and $\mathscr{P}(\mathscr{R}(R^{(1)}(a_{B}|x),...,R^{(m)}(a_{B}|x)))\leq \mathscr{P}(\mathscr{R}(R^{(1)}(a_{N}|x),...,R^{(m)}(a_{N}|x)))$, decide $x\in CTA(U)$;

(N1) If $\mathscr{P}(\mathscr{R}(R^{(1)}(a_{N}|x),...,R^{(m)}(a_{N}|x)))\leq \mathscr{P}(\mathscr{R}(R^{(1)}(a_{P}|x),...,R^{(m)}(a_{P}|x)))$  and $\mathscr{P}(\mathscr{R}(R^{(1)}(a_{N}|x),...,R^{(m)}(a_{N}|x)))\leq \mathscr{P}(\mathscr{R}(R^{(1)}(a_{B}|x),...,R^{(m)}(a_{B}|x)))$, decide $x\in NEA(U)$.
\end{theorem}

\noindent\textbf{Proof:} The proof is straightforward by Bayesian minimum risk theory.$\Box$

\begin{example}(Continuation from Example 5.8) First, for $x_{j}\in U$, by Theorem 5.2, we have
\begin{eqnarray*}
&&\mathscr{P}(\mathscr{R}(R^{(1)}(a_{P}|x_{j}),...,R^{(m)}(a_{P}|x_{j})))\\&&=\mathscr{P}(P(\Sigma^{m}_{i=1}k_{i}\ast\sqrt{1-(1-\mu^{2}_{\lambda^{(i)}_{PP}})^{\mathscr{P}(\mathscr{R}(x_{j}))}\ast (1-\mu^{2}_{\lambda^{(i)}_{PN}})^{1-\mathscr{P}(\mathscr{R}(x_{j}))}},\Sigma^{m}_{i=1}k_{i}\ast(\nu_{\lambda^{(i)}_{PP}})^{\mathscr{P}(\mathscr{R}(x_{j}))}\ast (\nu_{\lambda^{i}_{PN}})^{1-\mathscr{P}(\mathscr{R}(x_{j}))}));\\
&&\mathscr{P}(\mathscr{R}(R^{(1)}(a_{B}|x_{j}),...,R^{(m)}(a_{B}|x_{j})))\\&&=\mathscr{P}(P(\Sigma^{m}_{i=1}k_{i}\ast\sqrt{1-(1-\mu^{2}_{\lambda^{(i)}_{BP}})^{\mathscr{P}(\mathscr{R}(x_{j}))}\ast (1-\mu^{2}_{\lambda^{(i)}_{BN}})^{1-\mathscr{P}(\mathscr{R}(x_{j}))}},\Sigma^{m}_{i=1}k_{i}\ast(\nu_{\lambda^{(i)}_{BP}})^{\mathscr{P}(\mathscr{R}(x_{j}))}\ast (\nu_{\lambda_{BN}})^{1-\mathscr{P}(\mathscr{R}(x_{j}))}));\\
&&\mathscr{P}(\mathscr{R}(R^{(1)}(a_{N}|x_{j}),...,R^{(m)}(a_{N}|x_{j})))\\&&=\mathscr{P}(P(\Sigma^{m}_{i=1}k_{i}\ast\sqrt{1-(1-\mu^{2}_{\lambda^{(i)}_{NP}})^{\mathscr{P}(\mathscr{R}(x_{j}))}\ast (1-\mu^{2}_{\lambda^{(i)}_{NN}})^{1-\mathscr{P}(\mathscr{R}(x_{j}))}},\Sigma^{m}_{i=1}k_{i}\ast(\nu_{\lambda^{(i)}_{NP}})^{\mathscr{P}(\mathscr{R}(x_{j}))}\ast (\nu_{\lambda^{(i)}_{NN}})^{1-\mathscr{P}(\mathscr{R}(x_{j}))})).
\end{eqnarray*}

Second, by Definition 5.9, we have the Group Closeness Matrix $\mathscr{P}(R(S))$ as follows:
\begin{table}[H]\renewcommand{\arraystretch}{1.1}
\caption{The Group Closeness Matrix $\mathscr{P}(R(S))$.}
 \tabcolsep0.6in
\begin{tabular}{c c c c }
\hline  Action & $P$ & $B$ & $N$\\
\hline
$x_{1}$ &0.4103  &  0.4785  &  0.6907\\
$x_{2}$ &0.6448  &  0.4918  &  0.5519\\
$x_{3}$ &0.6887  &  0.4950  &  0.4810\\
$x_{4}$ &0.6616  &  0.4930  &  0.5287\\
$x_{5}$ &0.6582  &  0.4927  &  0.5338\\
$x_{6}$ &0.6246  &  0.4903  &  0.5752\\
\hline
\end{tabular}
\end{table}

Third, by Theorem 5.10, we have
\begin{eqnarray*}
POA(U)&=&\{x_{1}\};
CTA(U)=\{x_{2},x_{4},x_{5},x_{6}\};
NEA(U)=\{x_{3}\}.
\end{eqnarray*}
\end{example}

\section{Conclusions}

In this paper, we have presented three types of positive, central, and negative alliances with different thresholds, and employed examples to illustrate how to construct the positive, central, and negative alliances. We have studied conflict analysis of Pythagorean fuzzy information systems based on Bayesian minimum risk theory. Finally, we have investigated group conflict analysis of Pythagorean fuzzy information systems based on Bayesian minimum risk theory.

In practice, there are many types of dynamic Pythagorean fuzzy information systems such as with variations of object sets, attribute sets, and attribute value sets, and we will study other types of dynamic Pythagorean fuzzy information systems in the future.

\section*{Acknowledgments}

We would like to thank the reviewers very much for their
professional comments and valuable suggestions. This work is
supported by the National Natural Science Foundation of China (NO.61603063,
61273304,11526039), Doctoral Fund of Ministry of Education of China (No.201300721004), China Postdoctoral Science Foundation (NO.2015M580353), China Postdoctoral Science special Foundation (NO.2016T90383).

\end{document}